\documentclass{article}

\usepackage{natbib}
\usepackage[T1]{fontenc}

\usepackage{amsmath,amsfonts,bm}

\def\eqref#1{equation~\ref{#1}}

\def\1{\bm{1}}

\DeclareMathAlphabet{\mathsfit}{\encodingdefault}{\sfdefault}{m}{sl}
\SetMathAlphabet{\mathsfit}{bold}{\encodingdefault}{\sfdefault}{bx}{n}

\usepackage{graphicx}
\usepackage{subcaption}
\usepackage{wrapfig}
\usepackage{lipsum} 
\usepackage{hyperref}
\usepackage[capitalize,noabbrev]{cleveref}
\usepackage{url}
\usepackage{arydshln}
\usepackage{amsmath,amsthm}
\usepackage{graphicx} 
\usepackage{tabularx}
\usepackage{booktabs}
\usepackage{xspace}

\usepackage{adjustbox}
\usepackage[most]{tcolorbox}
\usepackage[svgnames]{xcolor}
\usepackage{nicematrix}
\usepackage{makecell} 
\usepackage{enumitem}
\usepackage{multicol,multirow}

\definecolor{pearDark}{HTML}{2980B9}
\newcommand{\iris}{{\text{IRIS}}\xspace}
\usepackage{listings}
\usepackage{siunitx}
\usepackage{tablefootnote}
\usepackage{threeparttable}
\usepackage[preprint]{icml2026}

\usepackage{amsmath}
\usepackage{amssymb}
\usepackage{mathtools}
\usepackage{amsthm}
\usepackage{listings}

\theoremstyle{plain}

\theoremstyle{definition}

\theoremstyle{remark}

\crefname{ineq}{inequ.}{inequ.}
\crefname{equation}{Eq.}{Eqs.}
\creflabelformat{ineq}{#2{\upshape(#1)}#3} 
\crefname{theorem}{Thm.}{Thm.}
\crefname{proposition}{Prop.}{Prop.}
\crefname{claim}{Claim}{Claims}
\crefname{algorithm}{Alg.}{Alg.}
\crefname{definition}{Def.}{Def.}
\crefname{lemma}{Lemma}{Lemmas}
\crefname{example}{Example}{Examples}
\crefname{appendix}{Appx.}{Appx.}
\crefname{figure}{Fig.}{Fig.}
\crefname{table}{Tab.}{Tab.}
\crefname{section}{Sec.}{Sec.}
\crefname{assumption}{Asm.}{Asm.}
\creflabelformat{ineq}{#2{\upshape(#1)}#3} 

\usepackage[textsize=tiny]{todonotes}

\icmltitlerunning{IRIS: Intrinsic Reward Image Synthesis}

\begin{document}

\twocolumn[
  \icmltitle{IRIS: Intrinsic Reward Image Synthesis}



  \icmlsetsymbol{equal}{*}

  \begin{icmlauthorlist}
    \icmlauthor{Yihang Chen}{equal,yyy}
    \icmlauthor{Yuanhao Ban}{equal,yyy}
    \icmlauthor{Yunqi Hong}{yyy}
    \icmlauthor{Cho-Jui Hsieh}{yyy}
  \end{icmlauthorlist}

  \icmlaffiliation{yyy}{Department of Computer Science, University of California, Los Angeles, USA}

  \icmlcorrespondingauthor{Yihang Chen}{yhangchen@cs.ucla.edu}

  \icmlkeywords{Machine Learning, ICML}

  \vskip 0.3in
]



\printAffiliationsAndNotice{}  

\definecolor{verylightgray}{gray}{0.875} 
\newcommand{\lightgray}[1]{\colorbox{verylightgray}{\strut #1}}

\begin{abstract}

Despite the success of Reinforcement Learning from Human Feedback (RLHF) in language reasoning, its application to autoregressive Text-to-Image (T2I) generation is often constrained by the limited availability of human preference data. This paper explores how an autoregressive T2I model can learn from internal signals without relying on external rewards or labeled data. Contrary to recent findings in math and code reasoning, we show that minimizing self-certainty, rather than maximizing it, improves image generation. We observe that autoregressive T2I models with higher certainty are likely to generate simple and uniform images, which are less aligned with human preferences, and models with lower certainty are likely to generate vivid images rich in detail. Based on this observation, we propose {\bf \iris} (\textbf{I}ntrinsic \textbf{R}eward \textbf{I}mage \textbf{S}ynthesis), the first framework to improve autoregressive T2I models with reinforcement learning using only an intrinsic reward. Empirical results demonstrate that applying IRIS to autoregressive T2I models achieves performance superior to those trained by individual external rewards, and matching those trained by ensemble external rewards. IRIS also incentivizes the emergence of nuanced CoT reasoning for high-quality image generation.
\end{abstract}
\vspace{-2em}
\section{Introduction} 
\label{sec:intro}
Reinforcement Learning (RL) has proven to be highly effective in advancing the reasoning capabilities of large language models, particularly in verifiable domains such as mathematics and programming~\citep{shao2024deepseekmath,hurst2024gpt,guo2025deepseek,hu2025open}. Motivated by these advances, a similar RL-based alignment approach is now being explored for text-to-image models~\citep{betker2023improving,team2023gemini,esser2024scaling}. However, applying RL here is more challenging due to the lack of verifiable rule-based rewards—unlike math or code, the quality of a visual output is inherently subjective and hard to evaluate automatically. { Existing methods in this area either build an image reward model from human preferences or use automated rewards from specialized models, such as object detectors~\citep{yan2024vigor} or Visual Question Answering (VQA) systems~\citep{jiang2025t2i}. However, the former approach is limited by the scalability and subjectivity of human labeling, while the latter is often domain-specific and struggles to generalize beyond the narrow settings for which it was trained. Besides, ~\citet{hong2026understanding} observe that reward usage is constrained to the specific training domain. For example, when trained with human preference reward HPSv2~\cite{wu2023human}, the model improves in overall aesthetics aligned with that reward, but fails to generalize to other abilities like spatial reasoning.

Therefore, the scalability, subjectivity and generalizability issues of external rewards call for better reward design for T2I models. Recently, several works have shown that in mathematical and code reasoning, the performance can be improved by maximizing the intrinsic reward by RL, i.e., the self-certainty~\citep{zhao2025learning,zhang2025right}. The intrinsic reward is particularly appealing for text-to-image generation, given the inherent difficulty of developing an explicit reward model. Motivated by this, we aim to answer the following question: 
\begin{quote} 
\textit{Is it possible to design a more general method for text-to-image generation using only intrinsic signals—without relying on human-labeled data or domain-specific heuristics, that is not constrained to one field?} \end{quote} 
}

\begin{figure*}[t]
  \centering
  \includegraphics[width=0.6\linewidth]{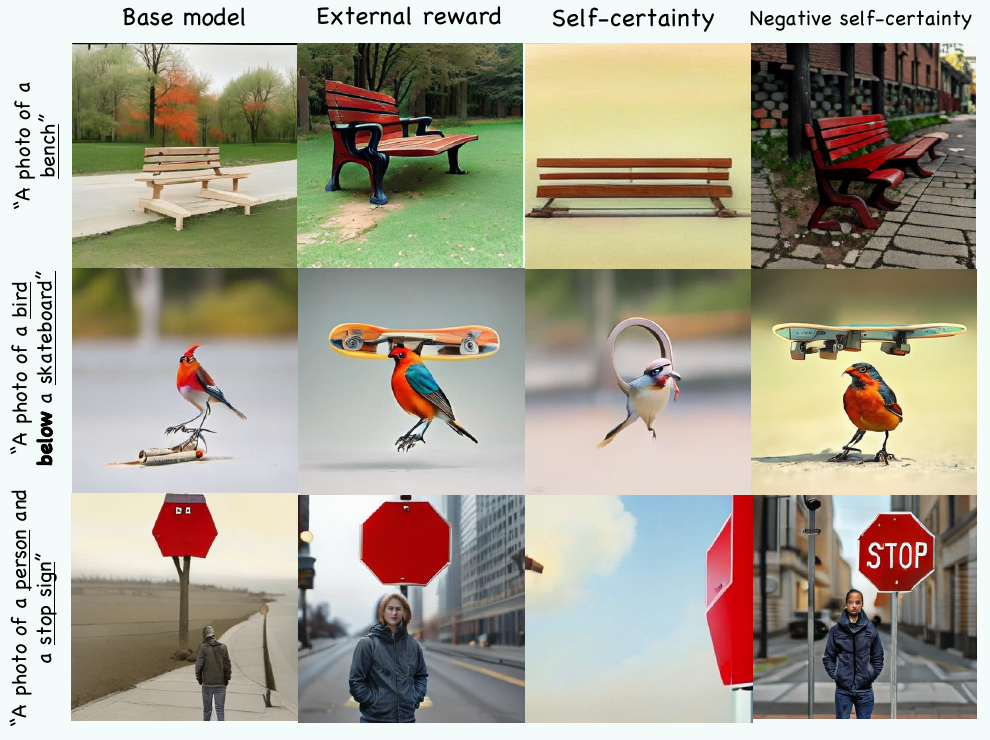}
  \caption{We perform RL fine-tuning on Janus-Pro using three reward schemes: (1) external reward (pretrained image reward models, etc), (2) self-certainty reward, and (3) negative self-certainty reward. The self-certainty is computed as the negative cross-entropy between the model's output distribution and a uniform distribution—where higher self-certainty indicates greater model self-certainty. The figure presents results across three tasks: (i) single-object generation, (ii) spatial generation, and (iii) two-object generation. We observe that increased self-certainty typically results in more uniform and less visually diverse images, while lower self-certainty tends to generate images with richer visual features that are more preferred by humans. Please refer to \cref{app:subsec:visualized_results} for more visualized results.}
  \label{fig:Intro}
      \vspace{-1.5em}
\end{figure*}

\begin{figure}[!htbp]
    \centering
    \includegraphics[width=0.7\linewidth]{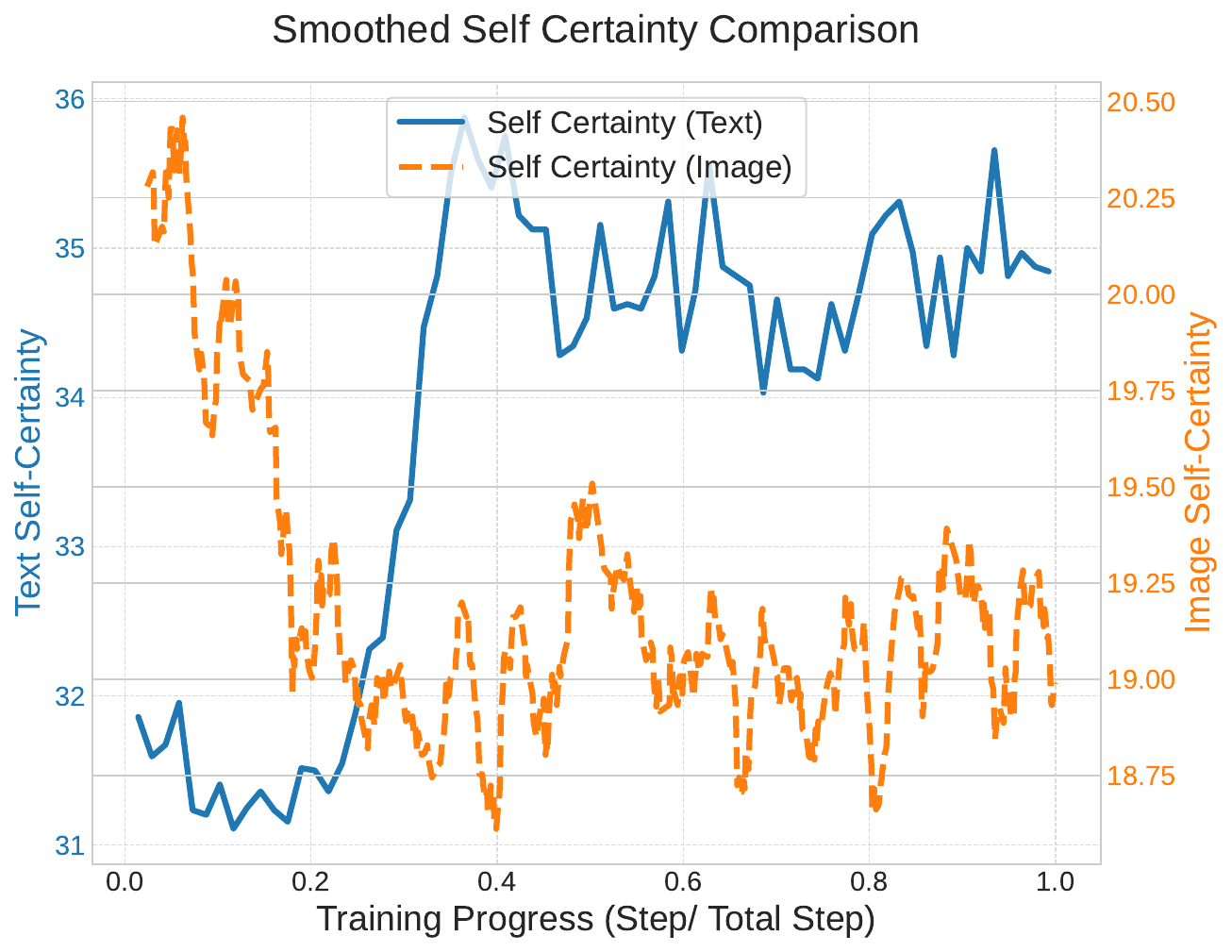}
    \captionof{figure}{{\bf Self-Certainty} on image tokens in the Janus-Pro-1B (orange line, right $y$-axis), and on text tokens in the Qwen2.5-1.5B-Instruct (blue line, left $y$-axis). {RL alignment with external reward models continuously increases the self-certainty of the LLM in the math reasoning, but decreases the self-certainty of the MLLM in the text-to-image generation.}}\label{fig:external_reward_sc}
    \vspace{-1em}
\end{figure}

We first analyze how an existing intrinsic signal, the Self-Certainty (SC)~\citep{zhao2025learning}, evolve in the LLM and MLLM's training. 
We compute the \textit{self-certainty} by the KL divergence between the uniform distribution and the model’s output distribution, measure how confident the model is in its predictions. We train an LLM and a text-to-image Multimodal LLM (MLLM), i.e., Qwen2.5-1.5B-Instruct on verified 0-1 accuracy reward in math problem, and Janus-Pro-1B on the average of HPSv2, GDino, GIT and ORM (see \cref{subsec:training_dertails} for details) by GRPO~\citep{shao2024deepseekmath} on text generation task and monitor the model's self-certainty on the text and image tokens respectively. 
Results in Fig~\ref{fig:external_reward_sc} show that as the training goes, the self-certainty of the LLM {in math reasoning tasks} continuously increases, but the self-certainty of the Janus-Pro model decreases. This indicates that external reward training makes the T2I model less self-confident in generating the images. We attribute this to the fact that when text-to-image models are highly confident in their generation, the outputs often exhibit a uniform and overly simplistic appearance, which are less preferred by human.

Based on this observation, we introduce {\bf \iris} (\textbf{I}ntrinsic \textbf{R}eward \textbf{I}mage \textbf{S}ynthesis), a RL framework that leverages the model’s the Negative Self-Certainty (NSC) as the reward signal. We note that \iris does not need any human knowledge or external verifiers, and is agnostic to the model architecture or training dataset. 

We find that \iris itself can significantly enhance the reasoning capabilities of T2I models and improving the image generation across diverse metrics. On Janus-Pro 1B models, \iris improves performance by 13.3\% on T2I-CompBench, 28.8\% on WISE, and 10.7\%/4.2\% on TIIF-short/long, showing that \iris improves reasoning and generation capabilities in the T2I models. 

Our contributions can be summarized in the following:
\vspace{-1em}
\begin{itemize}
\item We propose IRIS, the first RL-based alignment method to improve the T2I generation using only an intrinsic reward, where we \textbf{minimize the self-certainty} of the model.
\item {We observe that the model's self-certainty exhibits task-dependent behaviors: higher self-certainty benefits the model in domains requiring objective reasoning (e.g., math and code reasoning), but lower self-certainty benefits the model in subjective generation tasks (e.g., text-to-image generation). }
\item Empirically, we demonstrate that IRIS enhances T2I models without external supervision, \textbf{surpassing the performance by individual external rewards} and matching the performance by ensemble external rewards across multiple metrics. We attribute this to the fact that external rewards tend to constrain the model to a narrow domain, whereas \iris exploits the model’s intrinsic prior knowledge, enabling better generalization across diverse domains. IRIS also incentivizes the emergence of nuanced CoT reasoning for high-quality image generation.
\end{itemize}
\vspace{-2em}
\section{Related Work}
\paragraph{Reinforcement learning in image generation models} Reinforcement learning plays a key role in enhancing the performance of modern text-to-image generation models~\citep{betker2023improving,team2023gemini,wallace2024diffusion,esser2024scaling}. Early efforts primarily focused on training image reward models using human-labeled preference data~\citep{xu2023imagereward, wu2023human, xu2024visionreward}. To reduce reliance on manual annotation, subsequent work has explored automated reward generation. For example, \citet{yan2024vigor} leverage existing automatic scoring methods such as the Grounding DINO detector~\citep{liu2024grounding}. \citet{guo2025can} fine-tunes LLaVa-OneVision~\citep{li2024llava} to evaluate the alignment between the prompt and the generated image. Similarly, \citet{jiang2025t2i} utilize Visual Question Answering (VQA) models to provide feedback signals during training. However, our method get rid of the external models and guide the generation model with its intrinsic signals, making it adaptable to many scenarios. 
\vspace{-1em}

\paragraph{Reinforcement learning with intrinsic reward} Recent work in LLMs explores RLIF as a means to reduce reliance on human preference data or domain-specific verifiers. For instance, \citet{zhang2025right} and \citet{agarwal2025unreasonable} propose minimizing entropy as a form of reasoning incentivization. Building on this idea, \citet{zhao2025learning} introduce a self-certainty signal—defined as the cross-entropy between the output token distribution and a uniform distribution to guide RL training, reporting improved performance. In related efforts, \citet{prasad2024self} and \citet{zuo2025ttrl} generate multiple rollouts and leverage majority-vote outcomes to estimate advantages. To the best of our knowledge, we are the first to successfully train text-to-image generation models without external reward supervision. Our key insight lies in the observation that visual generative models exhibit lower self-certainty when producing visually rich and semantically meaningful images. This contrasts with findings in the LLM domain, where higher model self-certainty has been associated with better performance~\citep{zhang2025right,agarwal2025unreasonable,zhao2025learning}.

\vspace{-1em}

\section{Method}
\subsection{RL finetuning of LLMs} 
In RL finetuning, the LLM policy $\pi_\theta$ is optimized to maximize some reward function $r$. Given the input query $q$, the generated output $o$, the reference policy $\pi_{\rm ref}$, and the KL regularization coefficient $\beta$, the objective to optimize is
\begin{align}
\label{eq:rl_objective}
    \max_{\pi_\theta}\mathbb{E}_{\Vec{o}\sim \pi_\theta(\cdot|q)} \left[r(\Vec{o}|q)-\beta{\rm KL}(\pi_\theta(\Vec{o}|q))\|\pi_{\rm ref}(\Vec{o}|q)\right]\,. 
\end{align}
In this paper, we denote by $o_t$ the $t$-th token, and $\Vec{o}_{<t}$ the first $t-1$ tokens of the output $\Vec{o}$. We use $\operatorname{KL}(p\|q)$ to denote the KL divergence of the distribution $p$ and $q$. 

Currently, RL finetuning mainly consists of (1) reinforcement learning from human feedback (RLHF,~\citet{christiano2017deep, ouyang2022training,kaufmann2023survey}), where the reward function is typically learned explicitly or implicitly from human's preferences; (2) reinforcement learning from verifiable reward (RLVR,~\citet{lambert2023reinforcement}), where the reward function is verifiable, such as $0-1$ accuracy in mathematical problem solving and the object detector~\citep{liu2024grounding} in image generation; and (3) reinforcement learning from intrinsic reward (RLIF,~\citet{zhang2025right,agarwal2025unreasonable,zhao2025learning}), where the reward function is an intrinsic signal derived from the model’s internal state. 

However, existing RL finetuning methods mainly focus on text outputs, either in standard LLMs or vision language models. In this paper, we focus on RL finetuning of text-to-image generation with intrinsic feedback.

\subsection{\iris and reward design}
During training, given the prompt $q$, we first generate a semantic-level text description and then use it to guide the visual generation. We denote the output by $\Vec{o}$, which contains both text and image tokens. Conditioned on the prompt $q$ and the output before the $t$-th position $\Vec{o}_{<t}$, we define the Negative Self-Certainty (NSC) at position $t$ by the negative of Self-Certainty (SC,{~\citet{zhao2025learning}}):
\begin{align}
\label{eq:certainty}
    \operatorname{NSC}(o_t|q,\Vec{o}_{<t}) := -{\rm KL}(U\|\pi_{\theta}(o_{t}|q,\Vec{o}_{<t})), 
\end{align}
where $U$ is the uniform distribution on the vocabulary. 

Here, we the forward KL divergence instead, which encourages mode-covering behavior by rewarding probability distributions that cover multiple plausible outcomes. This stands in contrast to metrics like entropy (backward KL divergence), which are mode-seeking and favor a single high-probability output. Specifically, self-certainty mitigates the common bias against longer sequences found in perplexity and entropy-based measures, making it a more robust metric for a model's intrinsic self-certainty \citep{fang2024wrong,kang2025scalable}. Its practical value is supported by the recent work that it can serve as a powerful intrinsic reward to guide language model's learning across different domains \citep{zhao2025learning}. We also show in the ablation study of \cref{app:additional_ablation}, that forward KL outperforms backward KL. 

We maximize the objective \cref{eq:rl_objective} by applying Group-wise Relative Policy Optimization (GRPO,~\citet{shao2024deepseekmath}). GRPO's optimization process relies on sampling multiple candidates to inform policy updates. Specifically, for each query $q$, we generate a set of $G$ outputs $\{\Vec{o}_1, \ldots, \Vec{o}_G\}$ using a fixed behavior policy $\pi_{\theta_{\text{old}}}$. The relative rewards of these outputs are then used to estimate advantages, guiding the update for the target policy $\pi_\theta$ by maximizing the following objective:
\begin{equation}
\resizebox{\linewidth}{!}{$
\begin{aligned}
    & \mathcal{J}_{\text{GRPO}}(\theta) =  \mathbb{E}_{q \sim P(Q), O=\{\Vec{o}_i\}_{i=1}^G \sim \pi_{\theta_{\text{old}}}(\cdot|q)} \\ \nonumber
&   \frac{1}{G}\sum_{i=1}^G \frac{1}{|\Vec{o}_i|}\sum_{t=1}^{|\Vec{o}_i|}\Big\{
\min\left(c_{i,t}(\theta) \hat{A}_{i,t},\;\mathrm{clip}\big(c_{i,t}(\theta), 1-\epsilon,1+\epsilon\big)\, \hat{A}_{i,t}\right) .\\
&\qquad\qquad-\beta\mathrm{KL}\bigl(\pi_\theta\Vert\pi_{\mathrm{ref}}\bigr)\Big\}\,,  \nonumber
\end{aligned}
$}    
\end{equation}
where ratios $c_{i,t}$ are defined by $c_{i,t}(\theta) = \frac{\pi_\theta(o_{i,t}\mid q,\Vec{o}_{i,<t})} {\pi_{\theta_{\mathrm{old}}}(o_{i,t}\mid q,\Vec{o}_{i,<t})}$, and the advantage can be estimated by
    \begin{align*}
    u_i &= \sum_{t} \operatorname{NSC}(o_{i,t}|q,\Vec{o}_{i,<t}),\\
        \hat{A}_{i,t} &= \frac{u_{i} - \text{mean}(\{u_{1}, u_{2}, \cdots, u_{G}\})} {\text{std}(\{u_{1}, u_{2}, \cdots, u_{G}\})}\,.
    \end{align*}
{In the autoregressive T2I models, the generated output $\Vec{o}_i$ consists of text and image tokens, denoted by $\Vec{o}_{i,\text{text}}$ and $\Vec{o}_{i,\text{img}}$ respectively. $\Vec{o}_i$ is simply a concatenation of $\Vec{o}_{i,\text{text}}$ and $\Vec{o}_{i,\text{img}}$, i.e., $\Vec{o}_i=(\Vec{o}_{i,\text{text}},\Vec{o}_{i,\text{img}})$. Therefore, our objective is to maximize the  NSC for both text and image tokens. We validate this choice in the ablation study (\cref{subsec:ablation}, \cref{fig:ablation_minmax_image,fig:ablation_minmax_text}).} This is in clear contrast to previous works on language models in math reasoning, which maximize SC~\citep{zhao2025learning,zhang2025right}.

Notably, maximizing NSC for text tokens appears to contradict our observation in \cref{fig:external_reward_sc}, where we found that when training on reasoning tasks the text SC will increase (thus NSC will decrease). We suggest this discrepancy arises because of the task difference: math reasoning requires precise thought generation. However, our T2I setting first generates descriptive and explorative text from a given prompt before synthesizing the image. This need for textual exploration is supported by recent work~\citep{team2025tongyi}, which observed that when training the information-seeking agent, its output entropy (and thus NSC) also increases.

\section{Experiments }

\subsection{Experimental configuration}
\label{subsec:training_dertails} 
In this work, we use the MLLM, Janus-Pro~\citep{chen2025janus}, for the text-to-image generation task due to their strong instruction-following capabilities. To train our intrinsic reward by RL, we use the 553 GenEval~\citep{ghosh2023geneval} instructions as the training datasets. 
The key hyperparameters include: a learning rate of $1 \times 10^{-6}$, a KL regularization hyperparameter $\beta=0.01$, an effective batch size of 8, a maximum prompt length of 512 tokens, and a maximum completion length of 1024 tokens. In GRPO's advantage computation, we generate 8 text strings per query and subsequent 1 image per text string. 

To comprehensively assess our model's capabilities, we evaluate it against three diverse benchmarks, each designed to test different aspects of text-to-image abilities. First, T2I-CompBench~\citep{huang2023t2icompbench} targets compositional understanding, specifically assessing the model's capacity for attribute binding (e.g., color, shape, texture) and its handling of both spatial and non-spatial relationships between objects. Second, WISE (World Knowledge-Informed Semantic Evaluation,~\citet{niu2025wise}) measures the model's ability to apply real-world knowledge, evaluating performance on prompts requiring cultural common sense, spatio-temporal reasoning, and an understanding of natural sciences. Lastly, TIIF-Bench~\citep{wei2025tiif} is a comprehensive, and difficulty-graded benchmark designed to assess modern T2I models' ability to follow complex textual instructions. For each prompt, this benchmark contains the \emph{short} and \emph{long} versions to more systematically evaluate the model's ability to follow short and long instructions. 
Collectively, these benchmarks provide a multi-faceted evaluation of our model's performance, ranging from basic object composition to complex, knowledge-based semantic interpretation. We give a more detailed description of the benchmarks in \cref{app:benchmark_details}. Following previous benchmark results, we round the scores to two decimal places in the WISE benchmark, and four decimal places in the T2I-CompBench and TIIF-Bench. 

We will use the four external reward models to train the model as the external reward baseline (T2I-R1, ~\citet{jiang2025t2i}). 
Trained from human aesthetic preferences, the Human Preference Model (HPSv2, \citet{wu2023human}) assesses the overall aesthetic appeal and visual quality from the human perspective. To evaluate compositional accuracy, we use the GroundingDINO (GDINO, \citet{liu2024grounding}) object detector to verify the existence, count, and spatial arrangement of specified objects. Complementing this, a Visual Question Answering model GIT~\citep{wang2022git} question the image to confirm specific attributes, such as color and texture. Finally, an Output Reward Model (ORM, \citet{guo2025can}), a fine-tuned Large Multimodal Model, provides a holistic judgment on the alignment between the prompt and the generated image. We give a more detailed description of the external rewards in \cref{app:external_reward}.

\begin{figure*}[!htbp]
  \centering
  \includegraphics[width=0.9\linewidth]{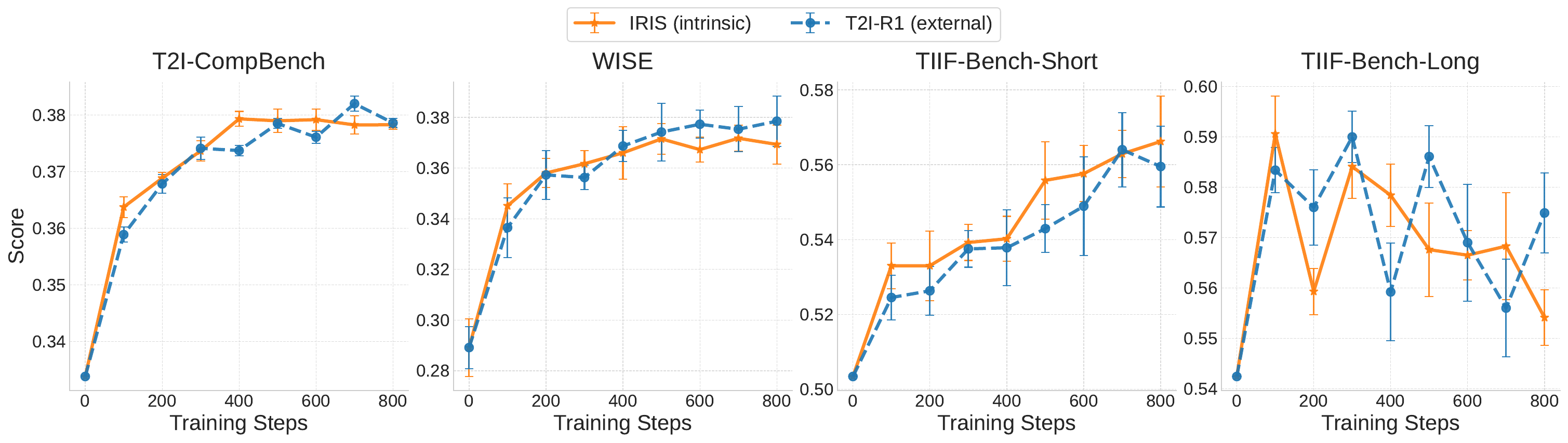}
  \caption{\textbf{Main results} of Janus-Pro 1B on T2I-CompBench, WISE and TIIF. {On T2I-CompBench and WISE, Our intrinsic reward \iris can achieve comparable results with external reward T2I-R1.}}
  \label{fig:main_bench}
\end{figure*}

\subsection{Main Results} 
\label{sec:main_results}
\paragraph{Performance on T2I benchmarks}
\cref{fig:main_bench} illustrates the averaged metrics across T2I-CompBench, WISE, and TIIF-Bench over the first 800 training steps. \cref{tab:benchmark_all} summarizes the 800-th step checkpoints compared against state-of-the-art baselines\footnote{{Baselines include the generation-only models: PixArt-$\alpha$~\citep{chen2023pixart}, SDXL~\citep{podell2023sdxl}, FLUX.1~\citep{flux2024,yang20241}, SD3-Medium~\citep{esser2024scalingrectifiedflowtransformers}; and the unified understanding and generation models: Show-o~\citep{xie2024show}, Show-o + PARM~\citep{guo2025can}, Orthus-7B~\citep{kou2024orthus}.}}. Detailed subscores over the first 800 steps are available in \cref{app:sec:subscores}.

Our results demonstrate that \iris significantly boosts the Janus-Pro 1B base model—improving performance by 13.3\% on T2I-CompBench, 28.8\% on WISE, and 10.7\%/4.2\% on TIIF-short/long, respectively. This performance is comparable to T2I-R1, which relies on ensemble external rewards. We observe two key trends: (1) The relative gain from RL fine-tuning is more pronounced on 1B models than 7B models, likely due to the stronger zero-shot capabilities of the larger base architectures. (2) \iris yields higher gains on short instructions (TIIF-short) compared to long ones. We attribute this to the emergence of descriptive CoTs; \iris incentivizes the model to "expand" short prompts into rich semantic representations. 

\paragraph{IRIS surpasses individual external rewards} In the main experiment, we utilize an ensemble of four rewards—HPSv2, GDINO, GIT, and ORM—to train the T2I-R1 models. To isolate the effect of each external reward, we evaluate Janus-Pro models trained with individual rewards on WISE. The results in \cref{tab:individual_reward} demonstrate that \iris's overall score surpasses aesthetic-focused HPSv2 and detection-based GDino by a significant margin across both the 1B and 7B scales. 

Meanwhile, although specialized rewards like HPSv2 maintain a slight lead in the \textit{Aesthetic} metric (0.6115 vs. 0.5445 for 1B), \iris demonstrates superior performance in \textit{Realism} and competitive \textit{Consistency}. This suggests that while external rewards are highly effective at optimizing the specific narrow distributions they were trained on, they often do so at the expense of other dimensions. For instance, HPSv2 significantly degrades the \textit{Realism} of Janus-Pro-7B (0.4745) compared to the base model (0.5470), whereas \iris improves it to 0.5710. This highlights a key advantage of intrinsic rewards: by incentivizing the model's own internal reasoning via maximizing NSC, \iris fosters a more balanced improvement across multiple facets of image generation without the ``reward hacking'' or distribution shifts often introduced by narrow external rewards.

\begin{table}[!htbp]
    \centering
        \caption{{\bf IRIS v.s. individual external rewards} on WISE benchmark: We train the base model with individual external reward as baselines. WiScore = 0.7 × Consistency + 0.2 × Realism + 0.1 × Aesthetic by the default setting.}
    \resizebox{\linewidth}{!}{
   \begin{NiceTabular}[color-inside]{lcccc}
    \toprule
      Model & Consistency$\uparrow$ & Realism$\uparrow$ & Aesthetic$\uparrow$ & WiScore$\uparrow$\\
      \midrule
        Janus-Pro-1B &$ 0.2420 $&$ 0.3550 $&$ 0.3790 $&$  0.2783 $\\
        +HPSv2 &$ 0.2960 $&$ 0.4515 $&$0.6115$&$ 0.3587 $\\
        +GDino &$ 0.2905 $&$0.4100$&$ 0.4395$&$ 0.3293$ \\
        +ORM &$ 0.3210  $&$ 0.4750 $&$ 0.5060 $&$ 0.3703 $\\
        +\textbf{IRIS} \rowcolor{verylightgray}&$ 0.3180  $&$ 0.4920 $&$ 0.5445 $&$ 0.3755 $\\
        \midrule
        Janus-Pro-7B &$ 0.4340  $&$0.5470 $&$0.5528$&$ 0.4685$ \\
        +HPSv2 &$ 0.4160 $&$0.4745$&$ 0.6530$&$ 0.4514$ \\
        +GDino &$ 0.4445$&$ 0.5170 $&$0.5435$&$ 0.4689$ \\
        +ORM &$ 0.4420 $&$ 0.5800$&$ 0.5975$&$ 0.4852$ \\
        +\textbf{IRIS}\rowcolor{verylightgray} & $0.4205$ & $0.5710$ & $0.6205$ & $0.4706$  \\
    \bottomrule
    \end{NiceTabular}}
    \label{tab:individual_reward}
\end{table}

\begin{table*}[!tbp]
\caption{{\bf IRIS v.s. ensemble T2I-R1}. ``Und.'' and ``Gen.'' denote ``understanding'' and ``generation'' respectively.  We report the scores of the last checkpoint of the T2I-R1 (external reward) and \iris (intrinsic reward). When evaluating the Janus Pro base, T2I-R1, and our \iris models, we will all use the semantic CoTs. We report the variance of the overall performance across 4 random seeds.  }
\centering
\begin{threeparttable}
\begin{subtable}{0.8\textwidth}
\centering
 \renewcommand{\arraystretch}{1.2}
 \vspace{-6pt}
\subcaption{{\bf T2I-CompBench}}
\vspace{-6pt}
\label{benchmark:t2icomp}
\begin{adjustbox}{width=\linewidth}
\begin{NiceTabular}[color-inside]
{clcccccc}
\toprule
\multicolumn{1}{c}
{\multirow{2}{*}{\bf Type}} &
\multicolumn{1}{c} 
{\multirow{2}{*}{\bf Method}} & \multicolumn{3}{c}{\bf Attribute Binding }$\uparrow$ & \multicolumn{2}{c}{\bf Object Relationship}$\uparrow$ & \multirow{2}{*}{\bf Overall$\uparrow$}
\\
\cmidrule(lr){3-5}\cmidrule(lr){6-7}
& &
{\bf Color } &
{\bf Shape} &
{\bf Texture} &
{\bf 2D-Spatial} &
{\bf Non-Spatial} &
\\
\midrule
\multirow{4}{*}{\rotatebox{90}{\it Gen. Only}} 
&PixArt-$\alpha$  & $ {0.6690} $ & {$ 0.4927 $} & $ {0.6477} $ & $ 0.2064 $ & $ {0.3197} $ & $ 0.3433 $\\
& SDXL& $0.5879$ &$0.4687$& $0.5299$& $0.2133$  & $0.3119$ & $0.3237$\\
&FLUX.1  & $ 0.7407 $ & $ {0.5718} $ & {$ 0.6922 $} & {$ 0.2863 $} & $ 0.3127 $ & $ {0.3703} $\\
& SD3-Medium &  $0.8132$ &$0.5885$ &$0.7334$ &$0.3200$&$0.3140$ &$0.3771$\\
\midrule
\multirow{8}{*}{\textit{\rotatebox{90}{Und. \& Gen.}}}&Show-o  & $ 0.56 $ & $0.41 $ & $0.46 $ & $0.20 $ & $0.30 $ & $0.29$ \\
&Show-o + PARM  & $ 0.75 $ & $0.56 $ & $0.66 $ & ${0.29} $ & $0.31 $ & $0.37$\\
\cdashline{2-8}
&Janus-Pro-1B &  $0.4922$& $0.2752$& $0.3965$& $0.1284$& $0.2964$& $0.3338$  \\
&T2I-R1-1B & $0.7973$& $0.4899$& $0.6659$& $0.3051$& $0.3079$& $0.3786_{\pm 0.0008}$
 \\
&\rowcolor{verylightgray}
\textbf{IRIS}-1B& $0.7902$& $0.5160$& $0.6983$& $0.2614$& $0.3059$& $0.3783_{\pm 0.0008}$
\\
&Janus-Pro-7B& $0.6518$& $0.4364$& $0.5529$& $0.1948$& $0.3097$& $0.3845$ \\
&T2I-R1-7B &  $0.8015$& $0.5661$& $0.7081$& $0.3246$& $0.3090$& $0.3992_{\pm 0.0019}$\\
&\rowcolor{verylightgray}
\textbf{IRIS}-7B & $0.7798$& $0.5130$& $0.6587$& $0.2468$& $0.3084$& $0.3906_{\pm 0.0022}$\\
\bottomrule

\end{NiceTabular}
\end{adjustbox}
\end{subtable}
\begin{subtable}{0.8\textwidth}
    \centering
 \renewcommand{\arraystretch}{1.2}
\subcaption{{\bf WISE}}
\vspace{-6pt}
\label{benchmark:wise}
\begin{adjustbox}{width=\linewidth}
\begin{NiceTabular}{clccccccc}[color-inside]
\toprule
\multicolumn{1}{c}{\multirow{2}{*}{\bf Type}} &
\multicolumn{1}{c}{\multirow{2}{*}{\bf Method}} &
\multirow{2}{*}{\bf Cultural$\uparrow$} &
\multicolumn{2}{c}{\bf Spatio-Temporal$\uparrow$} &
\multicolumn{3}{c}{\bf Natural Science$\uparrow$} &
\multirow{2}{*}{\bf Overall$\uparrow$} \\
\cmidrule(lr){4-5}\cmidrule(lr){6-8}
&&
{\bf } &
{\bf Time} &
{\bf Space} &
{\bf Biology} &
{\bf Physics} &
{\bf Chemistry} &
\\
\cmidrule{1-9}
\multirow{3}{*}{\rotatebox{90}{\it Gen. Only}} 
& PixArt-$\alpha$ & $0.45$ & $0.50$ & $0.48$ & $0.49$ & $0.56$ & $0.34$ & $0.47$ \\
& SD-XL & $0.43$ & $0.48$ & $0.47$ & $0.44$ & $0.45$ & $0.27$ & $0.43$ \\
& FLUX.1 & $0.48$ & $0.58$ & $0.62$ & $0.42$ & $0.51$ & $0.35$ & $0.50$ \\
\cmidrule{1-9}
\multirow{9}{*}{\textit{\rotatebox{90}{Und. \& Gen.}}} 
&Orthus-7B & 0.23 & 0.31 & 0.38 & 0.28 & 0.31 & 0.20 &  0.27\\
& Show-o & {$0.28$} & {$0.36$} & $0.40$ & $0.23$ & {$0.33$} & {$0.22$} & {$0.30$} \\
\cdashline{2-9}
&Janus-Pro-1B & $0.24$ & $0.28$ & $0.43$ & $0.28$ & $0.35$ & $0.15$ & $0.28$\\
&T2I-R1-1B  & $0.34$ & $0.42$ & $0.51$ &  $0.38$ & $0.43$ & $0.23$ & $0.38_{\pm 0.01}$\\
&\rowcolor{verylightgray}
{\bf IRIS}-1B   & $0.34$ & $0.41$ & $0.49$ &  $0.37$ & $0.41$ & {$0.22$} & $0.37_{\pm 0.01}$\\
&Janus-Pro-7B & $0.44$ & $0.49$ & $0.60$ & $0.45$ & $0.52$ & $0.27$ & $0.46$
\\
&T2I-R1-7B & $0.49$ & $0.51$ & $0.61$ & $0.48$ &  $0.53$ & $0.27$  & $0.50_{\pm 0.01}$ 
\\
&\rowcolor{verylightgray}
{\bf IRIS}-7B  &  $0.45$ & $0.51$ & $0.56$ & $0.44$ & $0.51$ & $0.28$  & $0.47_{\pm 0.01}$ 
\\
\bottomrule
\end{NiceTabular}
\end{adjustbox}
\end{subtable}
\vspace{-2em}
\begin{subtable}{0.8\textwidth}
    \centering
\small
\centering
 \renewcommand{\arraystretch}{1.2}
\subcaption{{\bf TIIF-Bench}}
\vspace{-6pt}
\label{benchmark:tiif}
\begin{adjustbox}{width=\linewidth}
\begin{NiceTabular}[color-inside]{
  c c l
  *{4}{c}              
  *{6}{c}              
  c                    
  c                    
}
\toprule
 &\multirow{2}{*}{\textbf{Type}} & \multirow{2}{*}{\textbf{Model}}
  & \multicolumn{4}{c}{\textbf{Basic Following}$\uparrow$}
  & \multicolumn{6}{c}{\textbf{Advanced Following}$\uparrow$}
  & \textbf{Designer$\uparrow$} 
  & \multirow{2}{*}{\textbf{Overall}$\uparrow$} \\ 

\cmidrule(lr){4-7} \cmidrule(lr){8-13} \cmidrule(l){14-14}
& & & Avg & Attr. & Rel. & Reas. & Avg & \makecell{Attr.\\+Rel.} & \makecell{Attr.\\+Reas.} & \makecell{Rel.\\+Reas.} & Style & Text & \makecell{Real\\World} & \\
\midrule
\multirow{11}{*}{\rotatebox{90}{\it Short Instructions}} &\multirow{4}{*}{\rotatebox{90}{\it Gen. Only}}
& PixArt-$\alpha$ & $0.5550$ & $0.5233$ & $0.6382$ & $0.5032$ & $0.3871$ & $0.3782$ & $0.5884$ & $0.4022$ & $0.5000$ & $0.0000$ & $0.4570$ & $0.4437$ \\
&& SDXL & $0.6572$ & $0.5933$ & $0.7757$ & $0.6032$ & $0.4973$ & $0.4782$ & $0.5622$ & $0.5259$ & $0.7333$ & $0.1683$ & $0.5092$ & $0.5496$ \\
&& FLUX.1 & $0.8312$ & $0.8705$ & $0.8725$ & $0.7501$ & $0.6579$ & $0.6707$ & $0.7384$ & $0.6909$ & $0.6667$ & $0.4383$ & $0.7072$ & $0.7109$ \\
&& SD3-Medium & $0.8020$ & $0.8450$ & $0.7890$ & $0.7721$ & $0.6233$ & $0.6654$ & $0.5792$ & $0.6146$ & $0.8000$ & $0.5339$ & $0.7164$ & $0.7017$ \\

\cmidrule{2-15}
&\multirow{7}{*}{\rotatebox{90}{\it Und. \& Gen.}}
& Show-o & $0.7308$ & $0.7483$ & $0.7882$ & $0.6557$ & $0.5367$ & $0.6095$ & $0.6859$ & $0.6646$ & $0.6333$ & $0.0383$ & $0.5502$ & $0.5972$ \\
&& Janus-Pro-1B & $0.6139$ & $0.7000$ & $0.5764$ & $0.5654$ & $0.4619$ & $0.5590$ & $0.4764$ & $0.3826$ & $0.5667$ & $0.0995$ & $0.6045$ & $0.5034$ \\
&& T2I-R1-1B  & $0.6943$ & $0.7300$ & $0.6690$ & $0.6839$ & $0.5276$ & $0.6207$ & $0.5493$ & $0.4675$ & $0.5000$ & $0.1176$ & $0.6978$  & $0.5595_{\pm 0.0096}$ \\
&& {\bf IRIS}-1B  \rowcolor{verylightgray} & $0.6799$ & $0.7700$ & $0.6982$ & $0.5715$ & $0.5262$ & $0.5619$ & $0.5533$ & $0.5000$ & $0.6000$ & $0.1584$ & $0.6828$ & $0.5662_{\pm 0.0132}$ \\
&& Janus-Pro-7B & $0.7321$ & $0.8100$ & $0.7014$ & $0.6848$ & $0.6098$ & $0.6327$ & $0.6172$ & $0.6235$ & $0.5333$ & $0.3348$ & $0.6418$ & $0.6199$ \\
&& T2I-R1-7B  & $0.7323$ & $0.7400$ & $0.7473$ & $0.7096$ & $0.6349$ & $0.6452$ & $0.6732$ & $0.6471$ & $0.4667$ & $0.3167$ & $0.6828$ & $0.6254_{\pm 0.0112}$ \\
&& {\bf IRIS}-7B  \rowcolor{verylightgray} & $0.7475$ & $0.7250$ & $0.7815$ & $0.7360$ & $0.6057$ & $0.6349$ & $0.6499$ & $0.5828$ & $0.5667$ & $0.2398$ & $0.7164$ & $0.6259_{\pm 0.0079}$ \\
\midrule
\multirow{11}{*}{\rotatebox{90}{\it Long Instructions}}&\multirow{4}{*}{\rotatebox{90}{\it Gen. Only}}& PixArt-$\alpha$ & $0.6100$ & $0.5633$ & $0.7407$ & $0.5257$ & $0.4490$ & $0.4132$ & $0.5246$ & $0.4709$ & $0.7667$ & $0.0083$ & $0.5316$ & $0.5050$ \\
&& SDXL & $0.5328$ & $0.5083$ & $0.6257$ & $0.4657$ & $0.3622$ & $0.3557$ & $0.4534$ & $0.3609$ & $0.6000$ & $0.0083$ & $0.4159$ & $0.4213$ \\
&& FLUX.1 & $0.7865$ & $0.8317$ & $0.8039$ & $0.7239$ & $0.6854$ & $0.7369$ & $0.7334$ & $0.7159$ & $0.6667$ & $0.5283$ & $0.7147$ & $0.7178$ \\
&& SD3-Medium & $0.7520$ & $0.7700$ & $0.7751$ & $0.7108$ & $0.6649$ & $0.7751$ & $0.6232$ & $0.6353$ & $0.7333$ & $0.2851$ & $0.6493$ & $0.6619$ \\
\cmidrule{2-15}
&\multirow{7}{*}{\rotatebox{90}{\it Und. \& Gen.}}& Show-o & $0.7583$ & $0.7983$ & $0.7832$ & $0.6932$ & $0.5038$ & $0.5682$ & $0.6896$ & $0.5622$ & $0.6667$ & $0.0283$ & $0.5092$ & $0.5886$ \\
&& Janus-Pro-1B & $0.6421$ & $0.7150$ & $0.6504$ & $0.5610$ & $0.5579$ & $0.5913$ & $0.6047$ & $0.5351$ & $0.5667$ & $0.0905$ & $0.5672$ & $0.5424$ \\
&& T2I-R1-1B  &  $0.6626$ & $0.7050$ & $0.7008$ & $0.5819$ & $0.5351$ & $0.5738$ & $0.5741$ & $0.5062$ & $0.5667$ & $0.1131$ & $0.6828$ & $0.5560_{\pm 0.0108}$ \\
&& {\bf IRIS}-1B \rowcolor{verylightgray} & $0.6689$ & $0.7100$ & $0.6792$ & $0.6175$ & $0.5420$ & $0.5981$ & $0.5606$ & $0.5111$ & $0.6333$ & $0.0995$ & $0.7052$ & $0.5683_{\pm 0.0121}$ \\
&& Janus-Pro-7B & $0.7459$ & $0.7550$ & $0.7874$ & $0.6952$ & $0.5781$ & $0.6273$ & $0.6117$ & $0.5524$ & $0.5000$ & $0.1991$ & $0.6903$ & $0.6020$ \\
&& T2I-R1-7B  & $0.7200$ & $0.7550$ & $0.7663$ & $0.6388$ & $0.6349$ & $0.6594$ & $0.6413$ & $0.6704$ & $0.5667$ & $0.1719$ & $0.6866$ & $0.6174_{\pm 0.0079}$ \\
&& {\bf IRIS}-7B \rowcolor{verylightgray} & $0.7549$ & $0.7550$ & $0.7641$ & $0.7456$ & $0.6011$ & $0.6022$ & $0.6178$ & $0.6255$ & $0.6333$ & $0.2308$ & $0.7201$ & $0.6327_{\pm 0.0055}$ \\
\bottomrule
\end{NiceTabular}
\label{tab:benchmark_tiif}
\end{adjustbox}
\end{subtable}
\end{threeparttable}
\label{tab:benchmark_all}
\end{table*}

\paragraph{IRIS incentivizes general T2I abilities matching ensemble external rewards} 
While ensemble rewards can mitigate the reward hacking in individual rewards, they are also limited by their own scope. We observe that IRIS is slightly behind ensemble-trained models in domains explicitly covered by external signals, such as basic aesthetics and object detection. However, IRIS demonstrates superior performance in specialized categories, i.e., the \emph{Designer: Real World} metric in TIIF-Bench.

This performance gap suggests that while ensembling multiple external rewards provides a safeguard against reward hacking, the resulting reward remains limited to the specific domains of those models. In ``unseen regimes'', such as styles not explicitly modeled by the ensemble, external rewards lose their efficacy. In contrast, IRIS incentivizes the model's inherent reasoning capabilities, allowing it to generalize to diverse, real-world prompts. We conclude that improving a model's latent abilities by intrinsic rewards enables more robust and generalizable exploration than relying exclusively on task-specific external signals.

\begin{figure}[!htbp]
\centering
\includegraphics[width=0.8\linewidth]{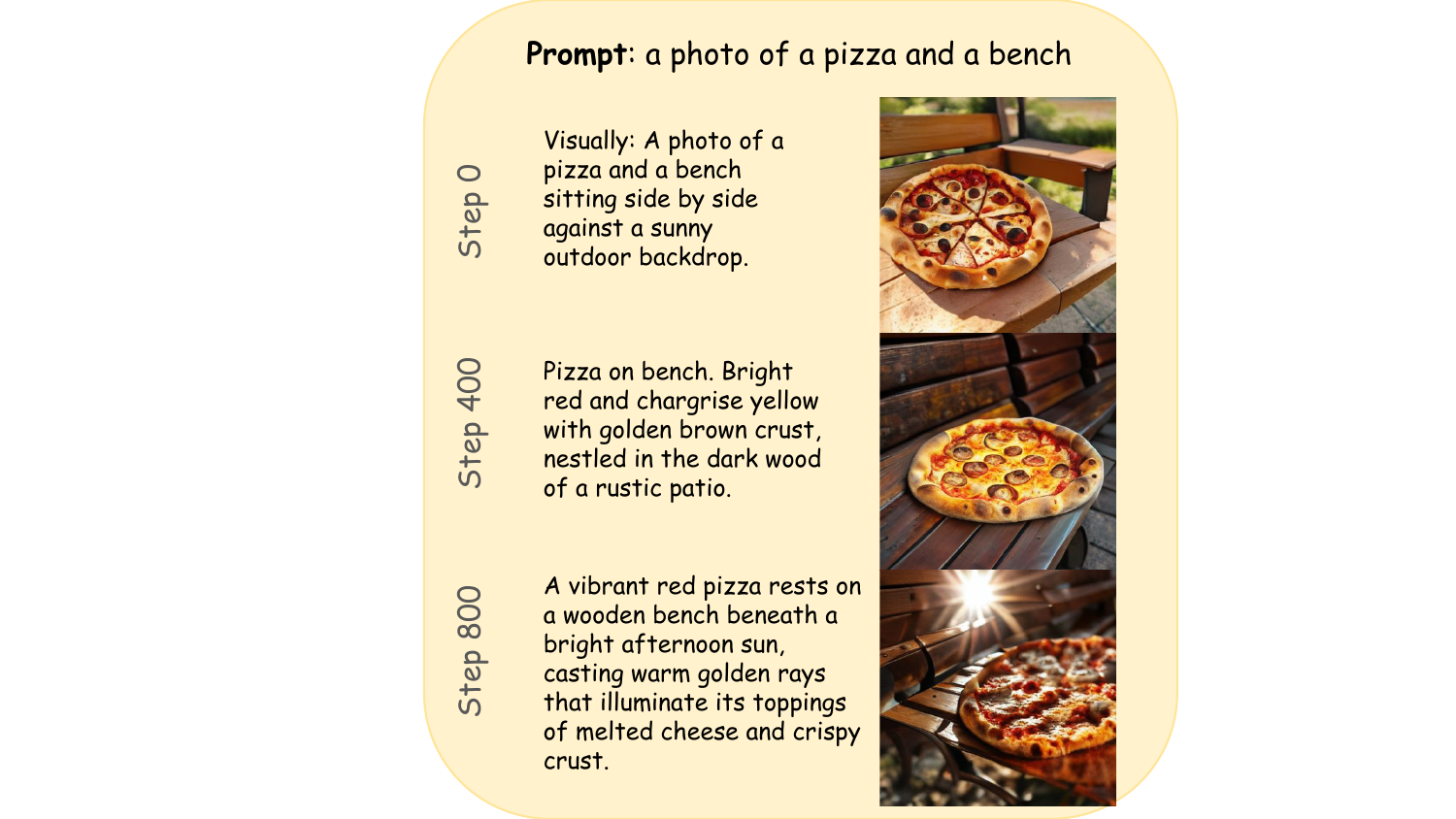}
\captionof{figure}{\textbf{Visualization of semantic CoTs in training}. We show that during the IRIS training, the generated CoTs will be increasingly more vivid, descriptive and rich in detail. 
}\label{fig:semantic_cot}
\vspace{-1.5em}
\end{figure}

\paragraph{IRIS incentivizes the emergence of descriptive CoT reasoning} 
Finally, we observe that intrinsic rewards guide the model toward generating semantically rich CoTs. As illustrated in \cref{fig:semantic_cot}, the CoTs generated during \iris training become progressively more vivid and detailed. This descriptive CoTs directly facilitates the diversity and accuracy of the final image. More examples are in \cref{app:subsec:failed_cases}.

\subsection{Ablation study}
\label{subsec:ablation}
\begin{figure}[!htbp]
  \centering
  \includegraphics[width=0.9\linewidth]{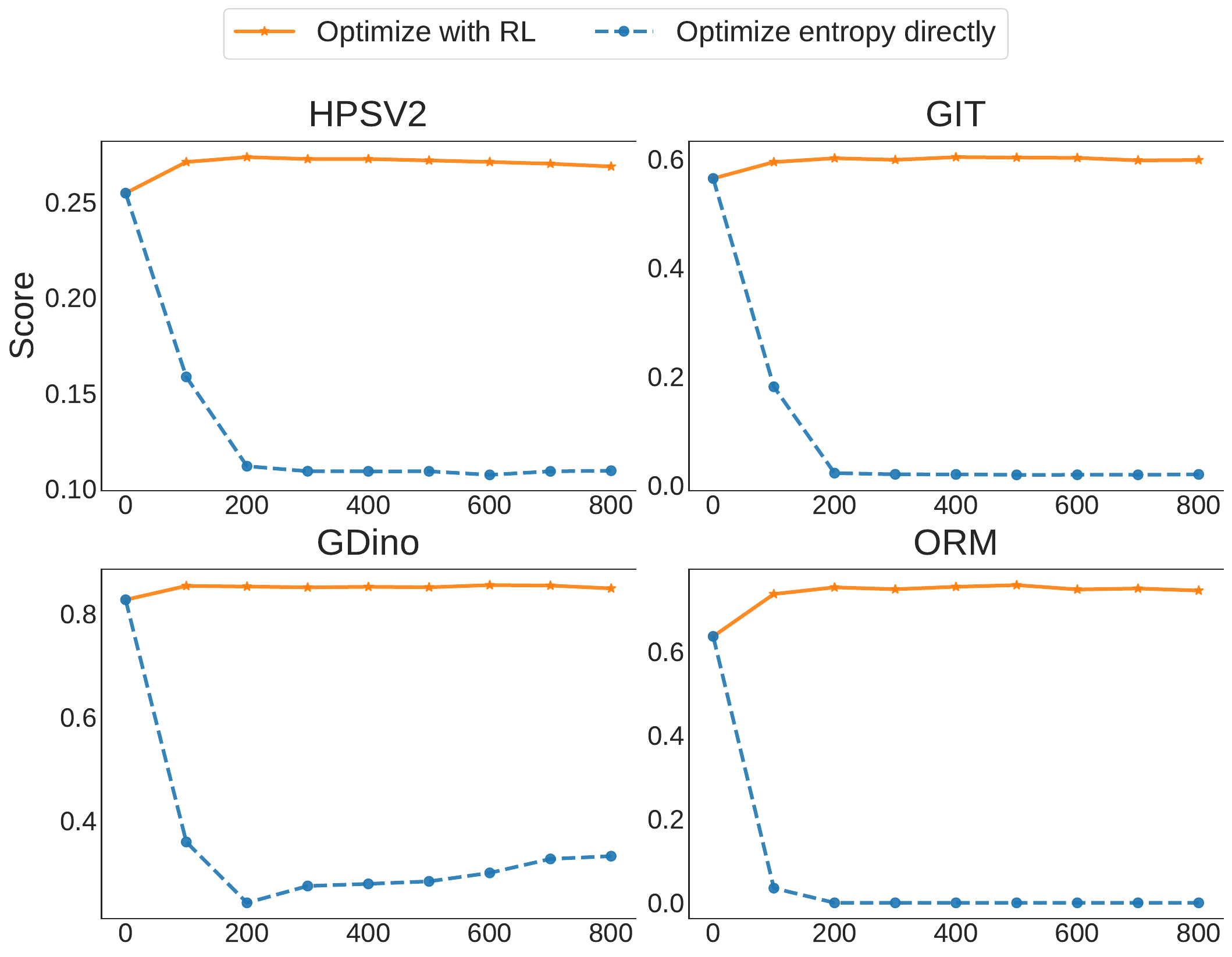} 
  \caption{\textbf{Ablation}: Optimizing with RL (ours) outperforms directly maximizing the NSC.}
  \label{fig:ablation_rl}
  \vspace{-1.5em}
\end{figure}

\paragraph{Evaluation metrics} We use four external rewards, namely HPSv2~\citep{wu2023human}, DINO~\citep{liu2024grounding}, ORM~\citep{guo2025can}, and GiT~\citep{wang2022git} introduced in \cref{subsec:training_dertails}, to evaluate the image generation in the ablation studies. 
Previously, we used these reward models to train the baseline T2I-R1. However, in our ablation studies on \iris, we never use these reward models in the training objectives, so they can be simple and unbiased metrics to evaluate the performance. We use 553 GenEval~\citep{ghosh2023geneval} instructions to synthesis the images. For each prompt, we generate four images and report averaged rewards to reduce noise in the ablation studies. 
\paragraph{RL or gradient descent}
We note that the intrinsic reward $\operatorname{NSC}(o_t|q,o_{<t})$ is differentiable w.r.t. the model parameters, and prior works on intrinsic reward~\citep{zhao2025learning,zhang2025right} all uses RL to optimize their differentiable intrinsic rewards. We ask if it is necessary to adopt RL rather than standard gradient descent to maximize the objective. Here, we maximize the following objective by gradient descent:
\begin{equation*}
\resizebox{\linewidth}{!}{$ \mathcal{J}_{\text{GD}}(\theta) =  \underset{\tiny\substack{
     q \sim P(Q),\\\Vec{o}_i \sim \pi_{\theta_{\text{old}}}(\cdot|q)}}{\mathbb{E}} 
     \frac{1}{G}\sum_{i=1}^G \frac{1}{|\Vec{o}_i|}\sum_{t=1}^{|\Vec{o}_i|}\left\{\operatorname{NSC}(o_{i,t}|q,\Vec{o}_{i,<t}) -\beta\mathrm{KL}\bigl(\pi_\theta\Vert\pi_{\mathrm{ref}}\bigr)\right\}\,,
$}
\end{equation*}
We adopt the same training configurations with the main experiments. 
In \cref{fig:ablation_rl}, we observe that directly maximizing the NSC will make the model generate meaningless images after hundreds steps of training. 

\begin{figure}[!htbp]
  \centering
  \includegraphics[width=0.9\linewidth]{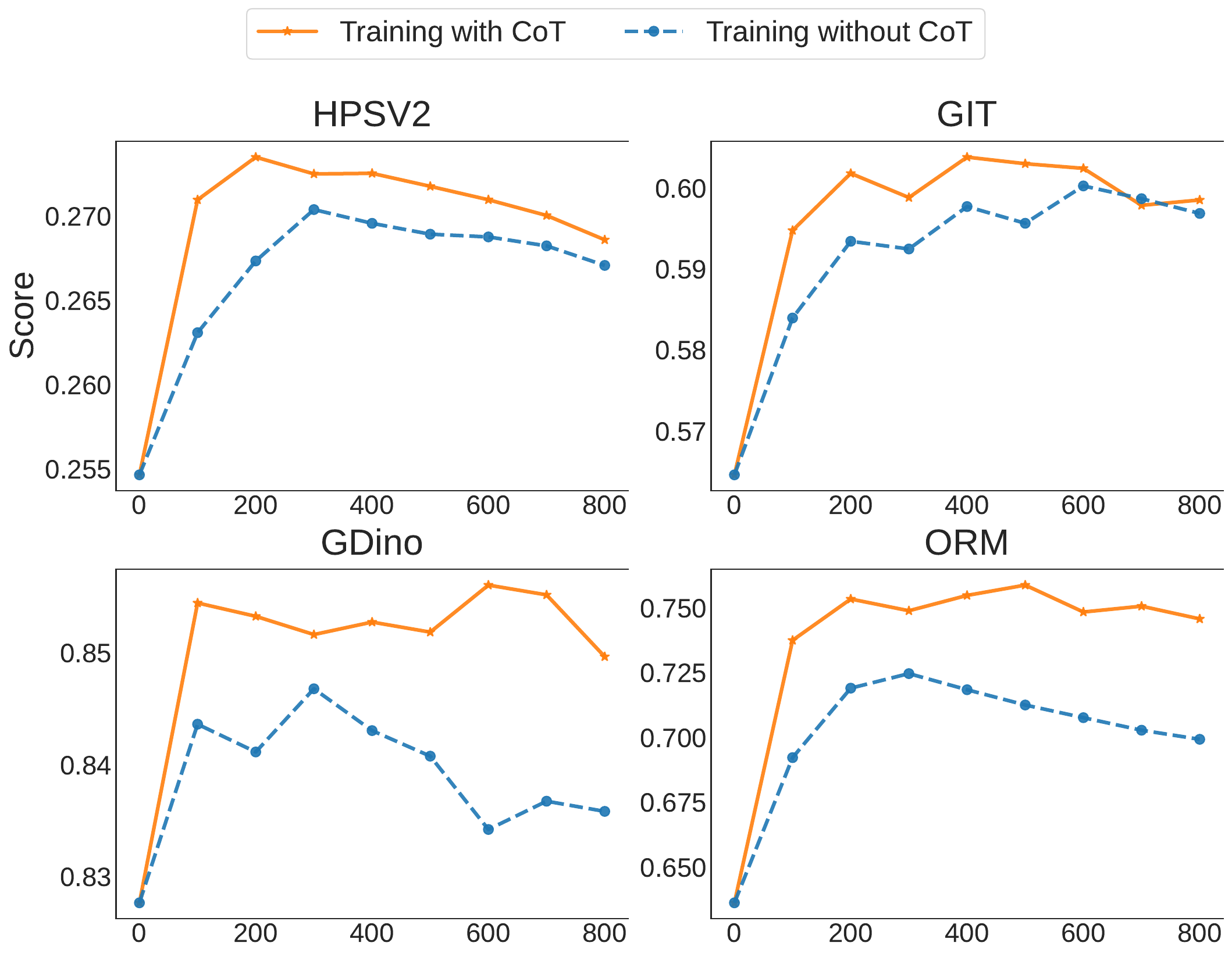} 
  \caption{\textbf{Ablation}: {training with CoTs (ours) outperforms without CoTs}. }
  \label{fig:ablation_cot}
  \vspace{-1.5em}
\end{figure}

\begin{figure}[!htbp]
  \centering
  \includegraphics[width=0.9\linewidth]{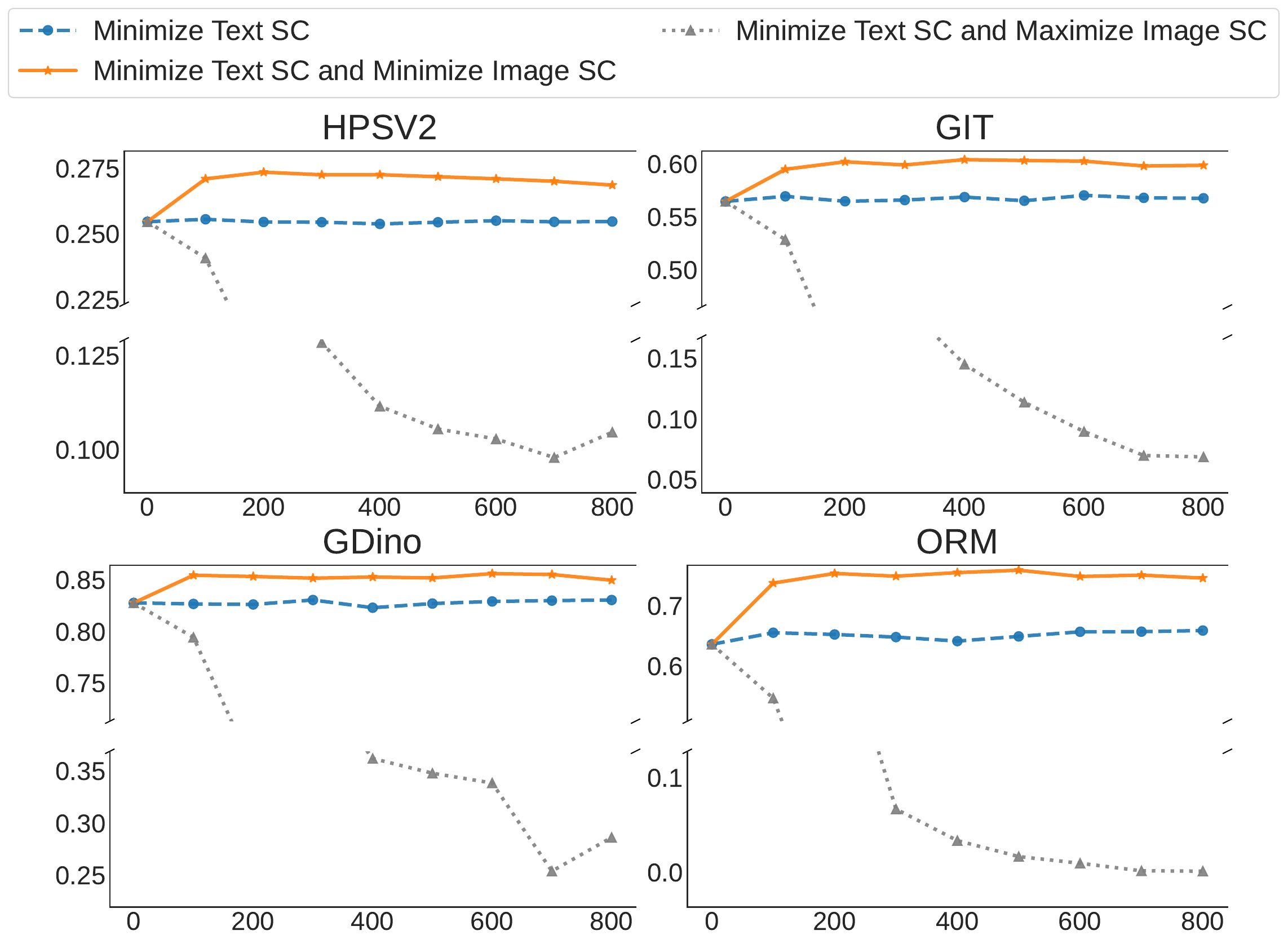} 
  \caption{\textbf{Ablation}: {minimizing image self-certainty (ours) outperforms maximizing it.}}
  \label{fig:ablation_minmax_image}
    \vspace{-1.5em}
\end{figure}

\begin{figure}[!htbp]
  \centering
  \includegraphics[width=0.9\linewidth]{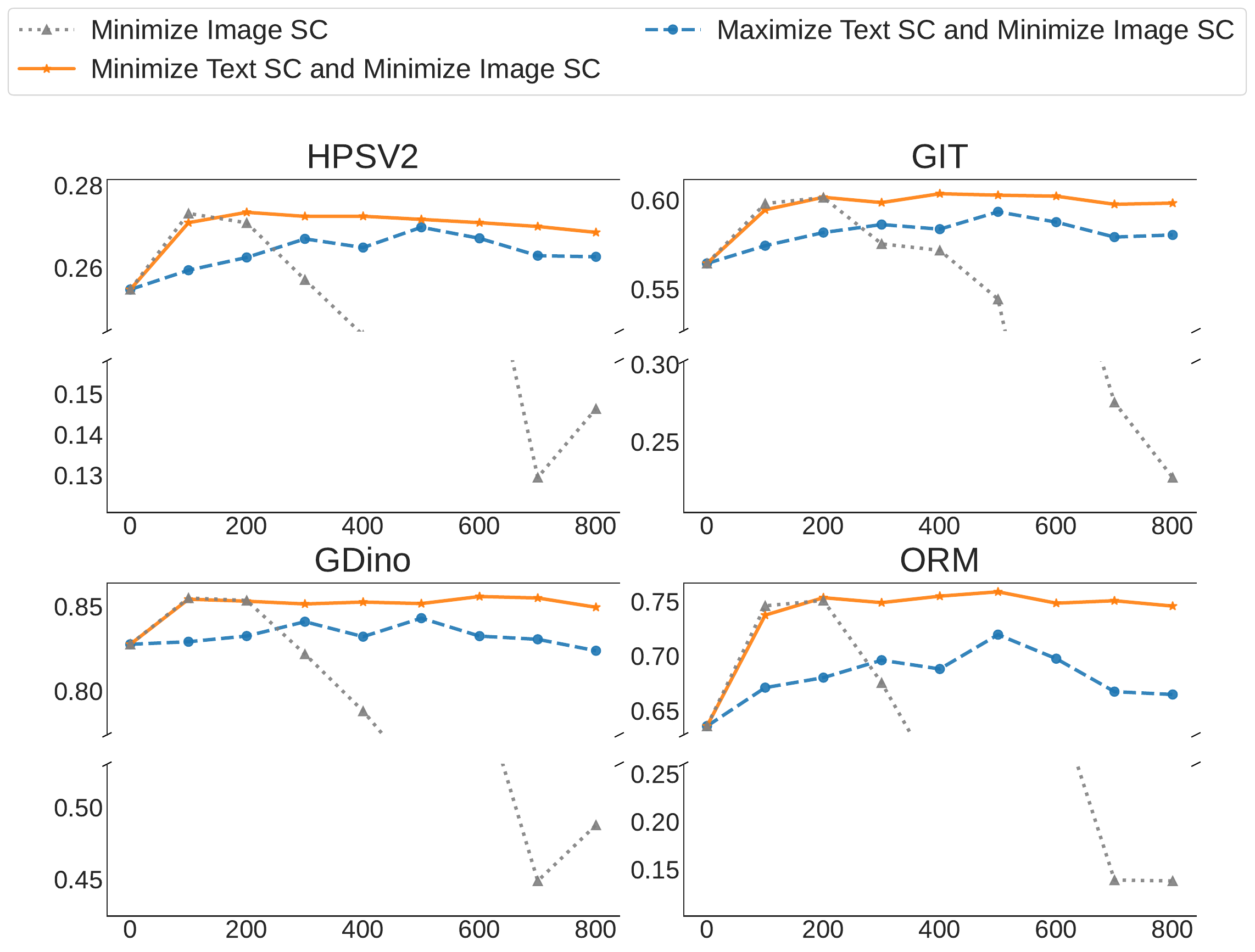}
  \caption{\textbf{Ablation}: minimizing text self-certainty (ours) outperforms maximizing it.}
  \label{fig:ablation_minmax_text}
    \vspace{-1.5em}
\end{figure}

We explain this phenomenon in the following: while fully random outputs maximize NSC, they lack semantic coherence. Consequently, the principle that 'increased NSC benefits T2I generation' does not imply that NSC should be maximized without bound. Our choice of the GRPO algorithm provides implicit regularization; by generating a batch of trajectories and aligning the model with those with the highest NSC, the optimization process naturally converges toward the most uncertain—yet still grounded—trajectories within that distribution.
\paragraph{Training with or without semantic CoTs}  
T2I-R1~\citep{jiang2025t2i} suggests training with semantic CoTs benefits training by external rewards. We show that training with semantic CoTs also benefits training on intrinsic rewards. 
We consider two series of models: (1) Janus-Pro trained by \iris, but without semantic CoTs, where we generate 8 images per query in GRPO's advantage computation. (2) Janus-Pro trained by \iris (with semantic CoTs), where we generate 8 text strings per query and subsequent 1 image per text string in GRPO's advantage computation. Results in ~\ref{fig:ablation_cot} show that training \iris with semantic CoTs consistently outperforms being without semantic CoTs. In \cref{fig:examples_compwise}, we present some generated figures of training with and without CoTs. Therefore, the default training and evaluation setting is with CoTs. 
\paragraph{Maximize or minimize image self-certainty} To determine whether image self-certainty should be maximized or minimized, we conduct three experiments: (1) minimizing text self-certainty only, (2) minimizing both text and image self-certainty (\iris), and (3) minimizing text self-certainty and maximizing image self-certainty. We run GRPO for 800 steps and evaluate four external rewards every 100 steps. In Figure~\ref{fig:ablation_minmax_image}, we show that minimizing both text and image self-certainty improves performance, but minimizing text self-certainty alone has little effect. Interestingly, maximizing image self-certainty actually degrades performance, causing a rapid drop, which supports our claim: higher self-certainty could hurt image generation.
\paragraph{Maximize or minimize text self-certainty} To evaluate whether text self-certainty should be maximized or minimized, we conduct three experiments: (1) minimizing image self-certainty only, (2) minimizing both text and image self-certainty (\iris), and (3) minimizing image self-certainty and maximizing text self-certainty. In Figure~\ref{fig:ablation_minmax_text}, we show that minimizing image self-certainty only achieves comparable performance in the early stages, however, it deteriorates rapidly after 200 steps. Meanwhile, minimizing text self-certainty always outperforms maximizing text self-certainty. This verifies our claim that maximizing text self-certainty discourages the model from exploring diverse semantic CoTs, thereby impairing its reasoning ability. In conclusion, minimizing text self-certainty proves to be a better strategy.
\subsection{Limitations and future works}
In this work, we examine our intrinsic reward strategy, \iris, on Janus-Pro, an autoregressive text-to-image model. While large language models have largely been dominated by the decoder-only architectures, the text-to-image models are far more diverse. They include a variety of competing model architectures such as continuous diffusion models~\citep{zhou2024transfusion}, masked-modeling approaches~\citep{xie2024show}, and MAE-style models~\citep{tong2024metamorph}, with no single architecture having dominated. Exploring how intrinsic reward can be adapted to these architectures is an interesting future research direction.
\vspace{-0.5em}
\section{Conclusion}
In this paper, we proposed \iris, which improves text-to-image models by leveraging negative self-certainty (NSC) as an intrinsic reward to maximize. Unlike RLHF or RLVF, our method doesn't need any human labeling or domain specific verifier, making it more scalable and easily generalizable to various domains. Empirical results demonstrate that applying IRIS to autoregressive T2I models achieves performance superior to those trained by individual external rewards and matching those trained by ensemble external rewards across diverse metrics. Our work highlights the different roles of self-certainty in different tasks, offering a potential guideline for the development of future multimodal generative models.

\newpage
\section*{Impact Statement}
Our work on text-to-image generation is committed to responsible AI development and adheres to standard academic and ethical practices. We recognize the potential for misuse of text-to-image generation models, including the generation of misleading or harmful content such as deepfakes. This particular project does not involve human subjects or raise concerns regarding data privacy, bias, or fairness in its current scope. Our research focuses on foundational architectural and training methodologies, with no direct application to the creation of sensitive or personally identifiable imagery. We are dedicated to ensuring that our research contributes to the safe and beneficial advancement of AI and are actively exploring methods to detect and prevent malicious applications of vision models.

\bibliographystyle{icml2026}
\bibliography{icml2026}

\newpage
\appendix
\onecolumn
\section{Benchmarks and reward models}
\label{app:bench_reward_details}
In this section, we give a detailed description of the benchmarks and external reward models we used in the main paper. 
\subsection{Benchmarks}\label{app:benchmark_details}

\paragraph{T2I-CompBench~\citep{huang2023t2icompbench}} T2I-CompBench is a compositional text-to-image generation framework for evaluating T2I models. For the attribute-binding task, we use disentangled BLIP-VQA~\citep{li2022blip} on three attributes: \textit{color}, \textit{shape} and \textit{texture}. For the object relationship, we use UniDet~\citep{zhou2022simple} for \textit{2D-spatial} relationship evaluation, and CLIPScore~\citep{radford2021learning} for \textit{non-spatial} relationship evaluation. In summary, we use the we a 3-in-1 evaluation metric \textit{complex} which computes the average score of CLIPScore, Disentangled BLIP-VQA, and UniDet, as the evaluation metric for complex compositions. Our evaluation is based on the 300 instructions in the T2I-CompBench's evaluation set of each attribute. For each instruction, our model generates four candidate images, and we report the averaged score of each attribute.

\paragraph{WISE~\citep{niu2025wise}} WISE (World Knowledge-Informed Semantic Evaluation) is a comprehensive benchmark designed to evaluate the ability of T2I models to integrate and apply real-world knowledge beyond merely word-to-pixel matching. It consists of 1,000 prompts spanning three main categories: \textit{cultural common sense}, \textit{spatio-temporal reasoning}, and \textit{natural science}. The spatio-temporal reasoning category is further divided into \textit{time} and \textit{space}, while natural science includes \textit{biology}, \textit{physics}, and \textit{chemistry}. WISE introduces WiScore, a scoring metric that uses GPT-4o to quantify \textit{consistency} (accuracy in depicting the prompt’s content), \textit{realism} (visual plausibility), and \textit{aesthetic quality} (composition and visual appeal). We report the average score of each category.

\begin{figure}[!ht]
    \centering
    \includegraphics[width=0.75\textwidth]{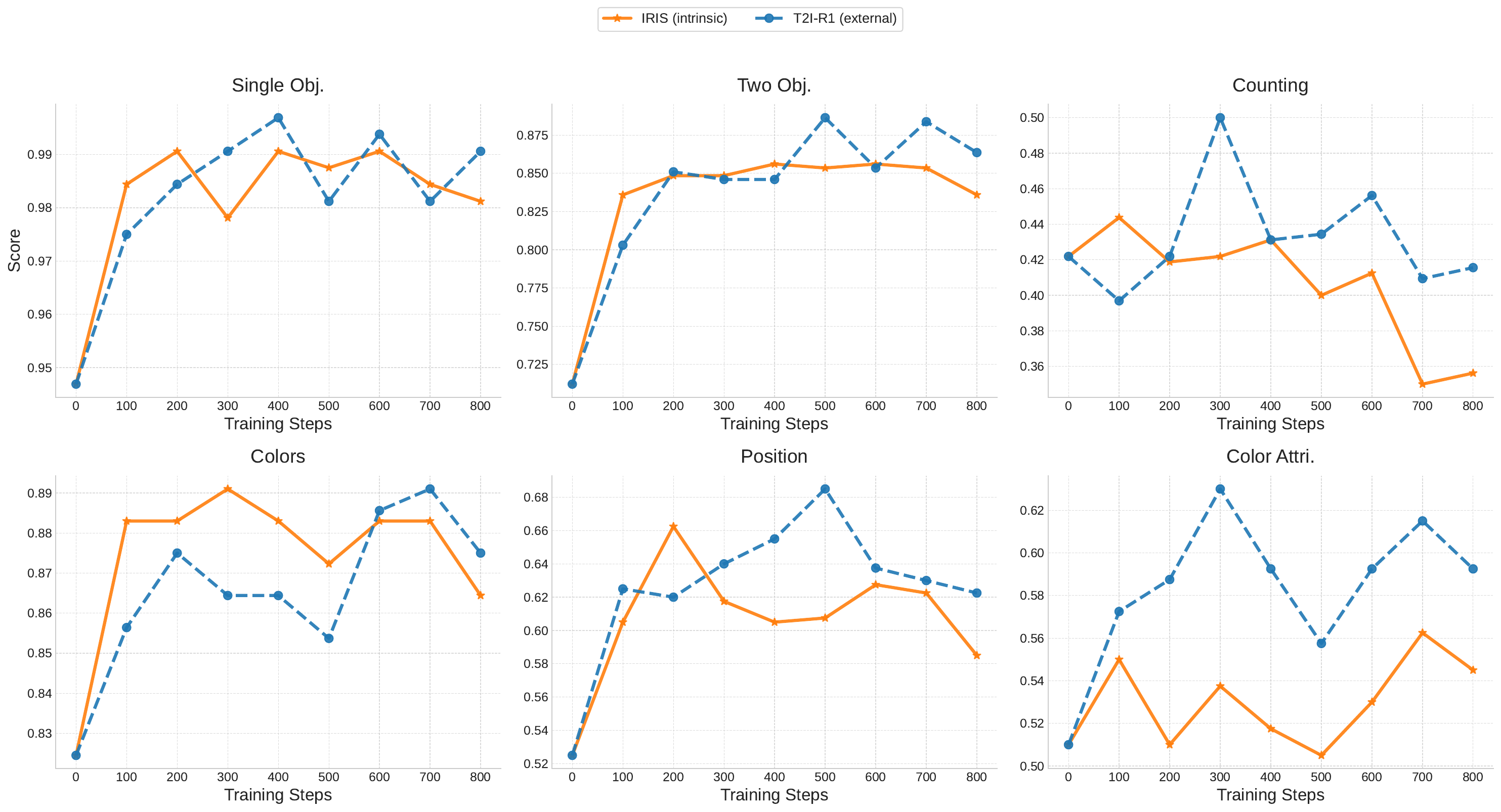}
    \caption{Sub-scores of {\bf GenEval} in the first 800 steps of training.}
    \label{fig:geneval_full}
\end{figure}

{\paragraph{TIIF-Bench~\citep{wei2025tiif}} TIIF-Bench (Text-to-Image Instruction Following Benchmark) is a comprehensive, and difficulty-graded benchmark designed to assess modern T2I models' ability to follow complex textual instructions, addressing limitations like simplistic prompts and coarse evaluation found in prior benchmarks. It has three major catgeries: \emph{Basic Following} (\emph{Attribute}, \emph{Relation}, \emph{Reasoning}), \emph{Advanced Following} (\emph{Attribute+Relation},\emph{Attribute+Reasoning}, \emph{Relation+Reasoning},\emph{Style},\emph{Text}), and \emph{Real World Following}. For each prompt, this benchmark contains the \emph{short} and \emph{long} versions to more systematically evaluate the model's ability to follow short and long instructions. 

\subsection{Reward models}
\label{app:external_reward}
\paragraph{Human Preference Model (HPSv2):}
To capture a generalized sense of image quality, we utilize a reward function derived from a Human Preference Model, HPSv2~\citep{wu2023human}. It is trained to learn human aesthetic preferences by learning from vast datasets of AI-generated images ranked by human annotators. When evaluating a new image, the model considers both its faithfulness to the text prompt and its overall visual appeal. These two factors are combined into a single score, which provides a comprehensive measure of the image's success. 

\paragraph{Object Detector (DINO):}
We employ an object detector, GroundingDINO~\citep{liu2024grounding} as a specialized "vision expert" to assess how accurately a generated image reflects the compositional elements of its prompt. This evaluation focuses on three key aspects: the existence of objects, their specified count, and their spatial relationships.

First, we parse the text prompt to create a target list of all mentioned objects $\{o_i\}_{i=1}^{K}$. The object detector then analyzes the generated image to locate these objects.
\begin{itemize}
    \item \textbf{Spatial relationships}: If a prompt describes a spatial arrangement (e.g., "a cup to the left of a book"), we use the detected bounding boxes of the objects. We then calculate metrics like their relative distance and Intersection over Union (IoU) to produce a spatial accuracy score, $\mathcal{R}_{\text{spatial}}$.
    \item \textbf{Object count}: Otherwise, if the prompt specifies a particular number of an object, $n_{o_i}$, (e.g., "three cats"), we compare this target to the number detected by the model, $\hat{n}_{o_i}$. 
    \item \textbf{Object existence}: Otherwise, for each of the $K$ target objects, we assign a binary score—$1$ if the object is detected in the image and $0$ if it is not. 
\end{itemize}
By combining these evaluations, the total reward from the object detector, $\mathcal{R}_{\text{Det}}$, is determined as:
\begin{equation*}
\renewcommand{\arraystretch}{1.5}
\begin{aligned}
{R}_{\text{Det}} =
\left\{
\begin{array}{ll}
\alpha \mathcal{R}_{\text{spatial}} + (1-\alpha) \frac{1}{K}\sum_{i=1}^{K} \mathbb{I}(o_i \text{ detected}), & \text{if spatial relationship in the prompt,} \\
\frac{1}{n}\sum_{i=1}^{K} \mathbb{I}(n_{o_i}=\hat{n}_{o_i}), & \text{if number in the prompt,} \\
\frac{1}{n}\sum_{i=1}^{K} \mathbb{I}(o_i \text{ detected}), & \text{else,} 
\end{array}
\right.
\end{aligned}
\end{equation*}
where $\mathcal{R}_{\text{spatial}}$ is 1 if the relative distance between the objects is larger than a threshold and the direction is right. If the direction is wrong, the reward is 0. Otherwise, we use the IoU as the spatial reward. We set $\alpha$ as 0.6 to encourage the correctness of the spatial relationship.

\paragraph{Visual Question Answering Model (GIT):}
We employ a Visual Question Answering (VQA) model, GIT~\citep{wang2022git}, to assess the presence and attributes of objects in generated images by answering image-related questions. The model is trained on question–answer pairs derived from visual content.

Our methodology involves transforming the image prompt into a series of targeted questions. For example, a prompt such as \textit{a blue bird and a red horse} is decomposed into individual queries like ``\textit{Is there a blue bird?}'' and ``\textit{Is there a red horse?}''. For each query $i$, we extract the model's output probabilities for the answers ``Yes'' ($P_{\text{Yes}}^{i}$) and ``No'' ($P_{\text{No}}^{i}$).

The final reward score, ${R}_{\text{VQA}}$, is computed by averaging the normalized probability of a ``Yes'' answer over all $K$ queries derived from the prompt. This is formally defined as:
$$
{R}_{\text{VQA}} = \frac{1}{K}\sum_{i=1}^{K}\frac{P^{i}_{\text{Yes}}}{P_{\text{Yes}}^{i} + P_{\text{No}}^{i}}.
$$
\paragraph{Output Reward Model (ORM):}
We incorporate an Output Reward Model (ORM) \citep{guo2025can} to provide an assessment of complete prompt-image alignment. The ORM is a Large Multimodal Model (LMM), such as LLaVA-OneVision~\citep{li2024llava-ov}, that has been specifically fine-tuned for this purpose. The fine-tuning objective instructs the model to act as a binary evaluator, outputting ``Yes'' only if the generated image perfectly aligns with the entire text prompt, and ``No'' otherwise.

The calculation of the reward, ${R}_{\text{ORM}}$, is similar to the VQA-based reward. The primary difference is that we provide the complete, original prompt to the ORM as a single query rather than decomposing it. The reward is thus the normalized probability of the model returning a ``Yes'' response for the complete prompt-image pair:
$$
{R}_{\text{ORM}} = \frac{P_{\text{Yes}}}{P_{\text{Yes}} + P_{\text{No}}}.
$$

\section{Additional experiments}\label{app:subscores}
\subsection{T2I-R1 implementation}
 We identify a key inconsistency in the official implementation of T2I-R1~\citep{jiang2025t2i}. Janus and Janus-Pro models use different chat templates: Janus models use keys "User" and "Assistant", but Janus-Pro models use keys "<|User|>" and "<|Assistant|>". \citet{jiang2025t2i} uses Janus model's chat template to train and evaluate the Janus-Pro models. In this paper, we will use the correct chat template to train and evaluate the Janus-Pro models, so numerical results reported in our paper are different from those in \citet{jiang2025t2i}. 




\subsection{Sub-category scores in benchmarks} \label{app:sec:subscores}
We give the sub-category score of three different benchmarks in \cref{fig:bencmark_full}. We discover that our intrinsic \iris is comparable to external rewards in the all the three benchmarks.
\begin{figure}[!htbp]
\centering
\begin{subfigure}{0.70\linewidth}
    \centering
    \includegraphics[width=\linewidth]{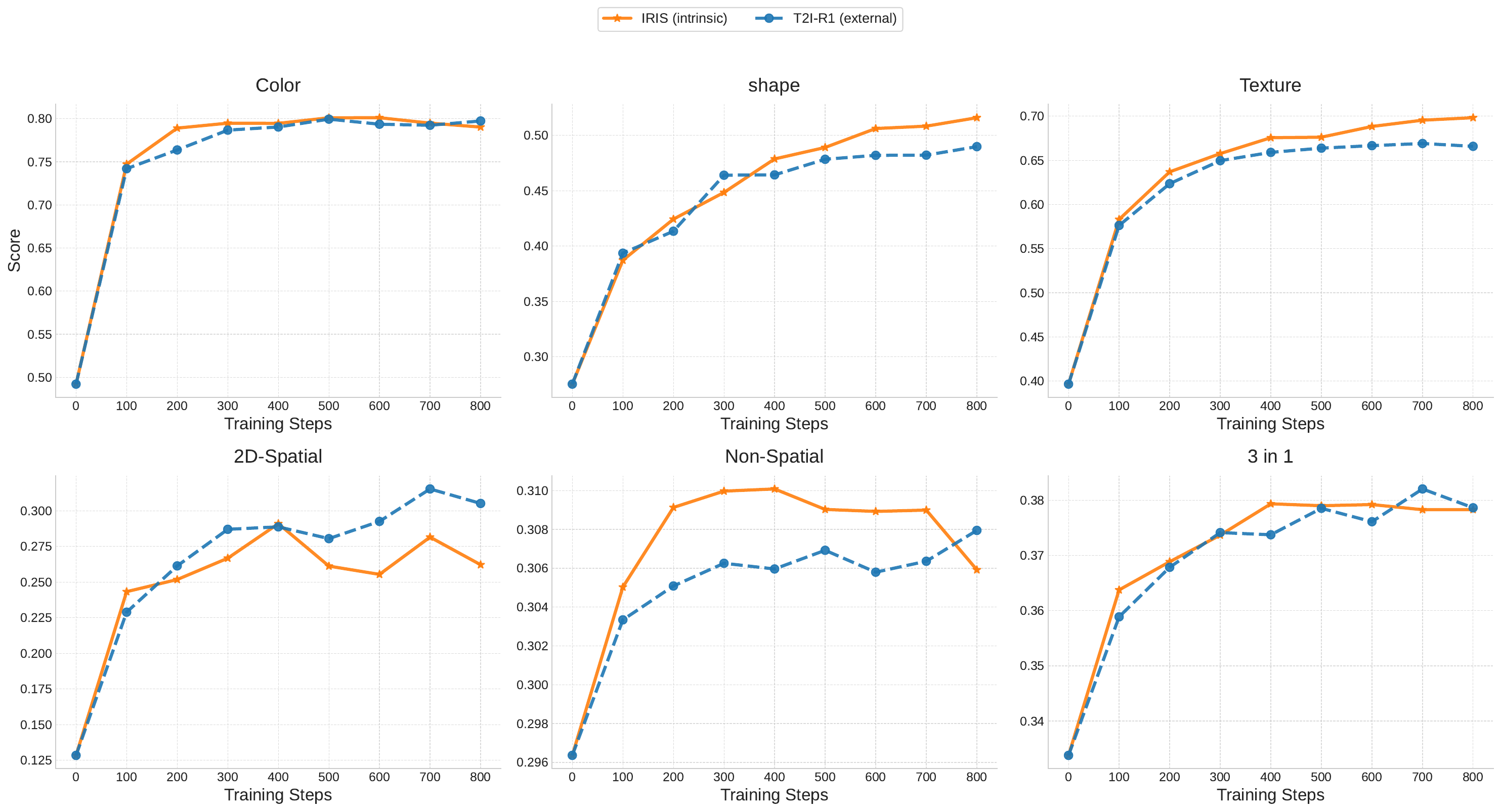}
    \subcaption{{\bf T2I-CompBench}}
    \label{fig:compbench_full}
\end{subfigure}
\begin{subfigure}{0.70\linewidth}
    \centering
    \includegraphics[width=\linewidth]{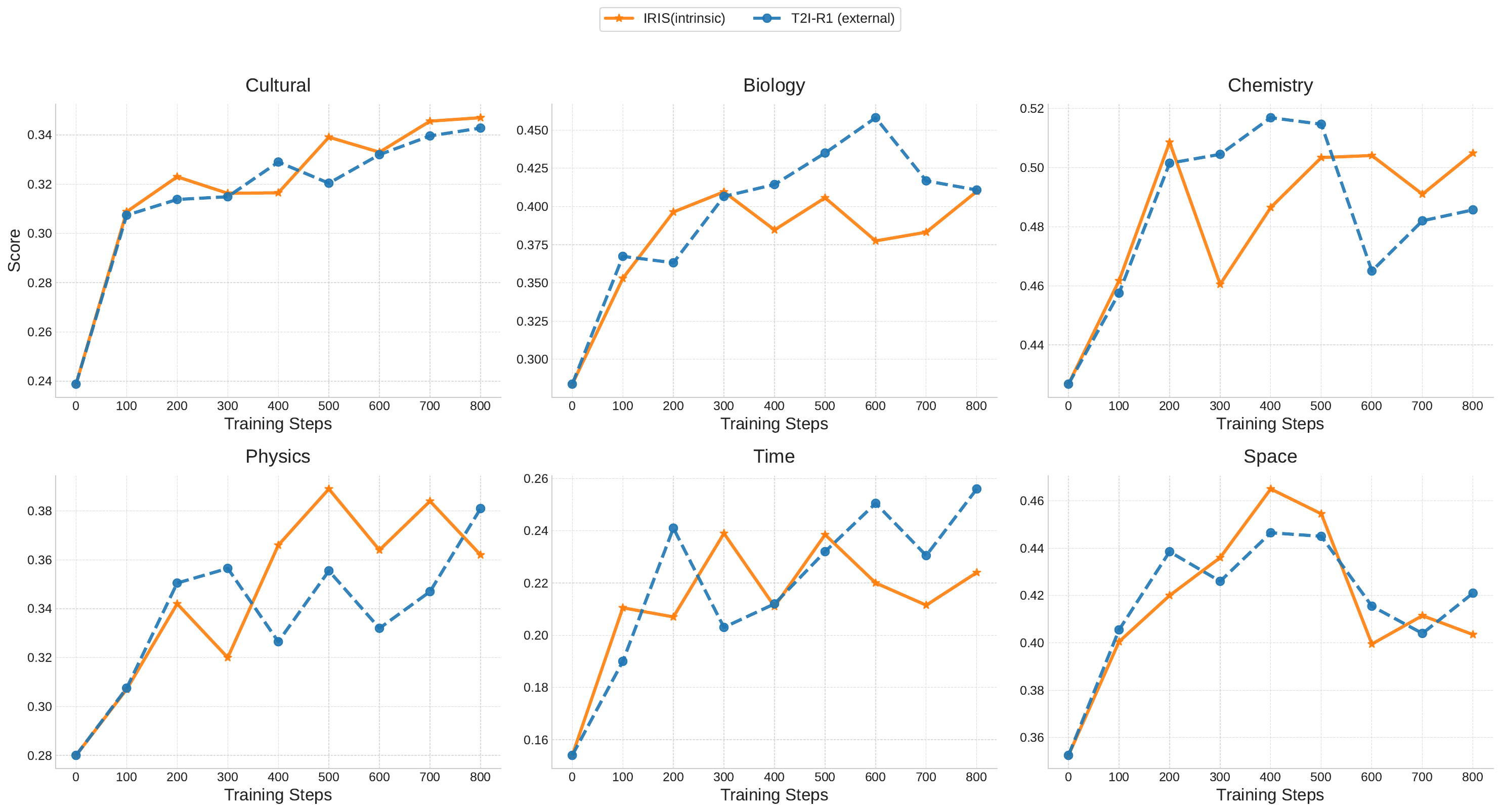}
    \subcaption{{\bf WISE}}
    \label{fig:wise_full}
\end{subfigure}
\begin{subfigure}{0.70\linewidth}
    \centering
    \includegraphics[width=\linewidth]{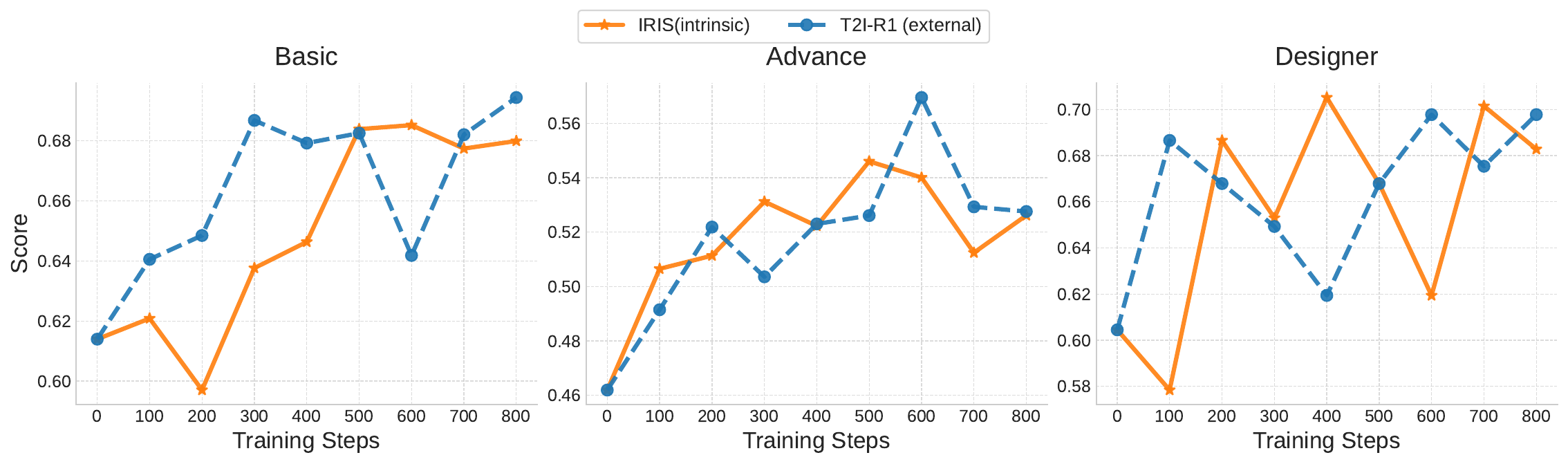}
    \subcaption{{\bf TIIF-Bench-Short}}
    \label{fig:tiif_short}
\end{subfigure}
\begin{subfigure}{0.70\linewidth}
    \centering
    \includegraphics[width=\linewidth]{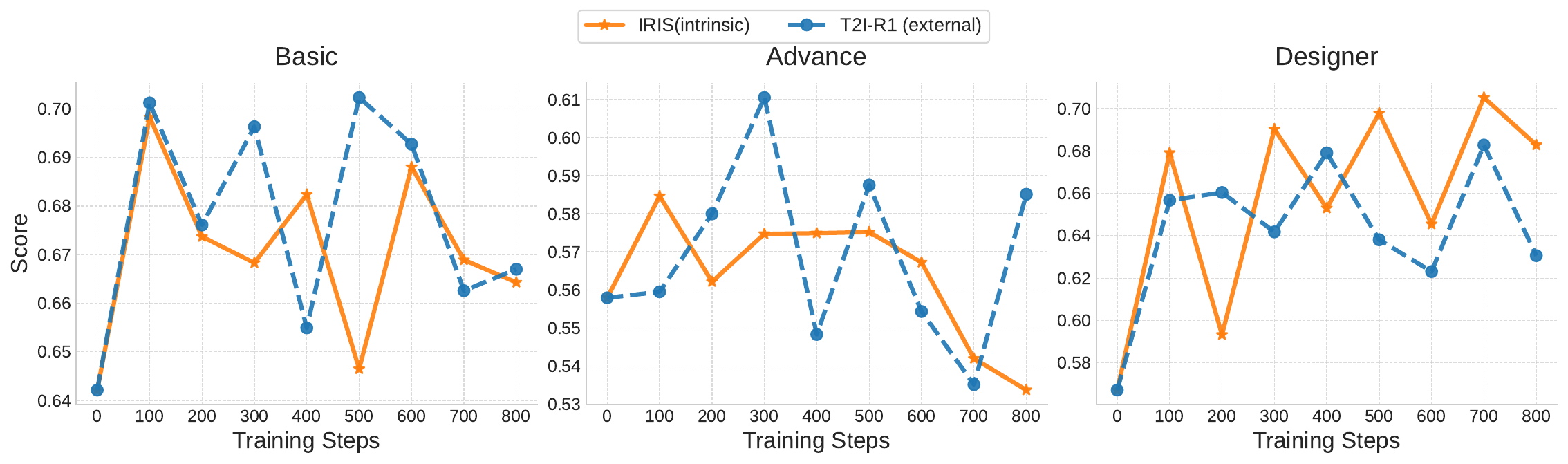}
    \subcaption{{\bf TIIF-Bench-Long}}
    \label{fig:tiif_long}
\end{subfigure}
\caption{Subscores of Janus-Pro-1B on T2I-CompBench, WISE, TIIF-Bench-Short and TIIF-Bench-Long }
\label{fig:bencmark_full}
\end{figure}


{\subsection{Human Evaluation} For the human evaluation, we randomly sample 100 prompts from the three benchmarks (WISE, T2I-Compbench, TIIF). Due to limited resources, we only evaluated three checkpoints: (1) \textbf{Janus-Pro-1B} (base generation model), (2) {T2I-R1 (external reward)}: Janus-Pro-1B finetuned with four external rewards, and (3) {IRIS (internal reward)}: Janus-Pro-1B finetuned with our intrinsic reward. For (2) and (3), we use the best checkpoints selected on WISE. We then ask 15 human evaluators to choose the best image among the three for each prompt and record the number of times each model is upvoted. The results are reported in \cref{tab:human_eval}. Our method achieves performance comparable to the external-reward-based model.}
\begin{table}[!htbp]
\centering
\begin{tabular}{lccc}
\hline
Method & Janus-Pro-1B        & T2I-R1 External reward & IRIS (Intrinsic reward) \\
\hline
Rate & 0.13                    & 0.42          &0.45         \\
\hline
\end{tabular}
\caption{{Human evaluation results}}\label{tab:human_eval}
\end{table}
\vspace{-1em}
\subsection{GenEval benchmark}
 GenEval~\citep{ghosh2023geneval} is a object-centric framework for evaluating T2I models. First, we use the model to generate images based on testing prompts, which are divided into 6 categories: (1) \textit{Single Object} (2) \textit{Two Objects} (3) \textit{Colors} (4) \textit{Counting} (5) \textit{Position} (6) \textit{Color Attribution}. After image generation, we use an object detector~\citep{chen2019mmdetection,cheng2022masked} to detect the targeted objects and CLIP ViT-L/14 to classify the object color. Each image receives a binary score indicating whether the described object is rendered correctly. Our evaluation is based on the 553 instructions in the GenEval's evaluation set. For each instruction, our model generates four candidate images, and we report the averaged score in one category.} Compared with the benchmarks T2I-CompBench, WISE, and TIIF-Benhch in the main paper, GenEval is an easier and less comprehensive benchmark. We mainly use its instructions for training and ablation studies, and use more up-to-date benchmarks for evaluation. For the completeness, we present the sub-scores in training in \cref{fig:geneval_full} and the benchmark results in \cref{tab:benchmark_geneval}  

\begin{table}[!htbp]
    \centering
    \renewcommand{\arraystretch}{1.2}
    \small
    \caption{\textbf{GenEval.} benchmark results.}
    \vspace{-2pt}
    \resizebox{0.8\linewidth}{!}{
    \begin{NiceTabular}{clccccccc}[color-inside]
        \toprule
        \textbf{Type} & \textbf{Method}  & \textbf{Single Obj.$\uparrow$} & \textbf{Two Obj.$\uparrow$} & \textbf{Counting$\uparrow$} & \textbf{Colors$\uparrow$} & \textbf{Position$\uparrow$} & \textbf{Color Attri.$\uparrow$} & \textbf{Overall$\uparrow$} \\
        \midrule
        \multirow{4}{*}{\textit{\rotatebox{90}{Gen. Only}}} 
        & PixArt-$\alpha$ &  ${0.98}$ & $0.50$ & $0.44$ & $0.80$ & $0.08$ & $0.07$ & $0.48$ \\
        & SDXL &  ${0.98}$ & $0.74$ & $0.39$ & $0.85$ & $0.15$ & $0.23$ & $0.55$ \\
        & FLUX.1  & $0.98$ &$0.81$ &{$0.74$}& $0.79$& $0.22$ & $0.45$ & $0.64$ \\
        & SD3-Medium & ${0.99}$ & ${0.94}$ & ${0.72}$ & ${0.89}$ & $0.33$ & ${0.60}$ & ${0.74}$ \\
        \midrule
        \multirow{7}{*}{\textit{\rotatebox{90}{Und. \& Gen.}}}
        & Show-o & ${0.98}$ & $0.80$  & ${0.66}$ & $0.84$ & $0.31$ & $0.50$ & $0.68$ \\
        \cdashline{2-9}
        &Janus-Pro-1B &  $0.94$& $0.71$& $0.42$& $0.82$& $0.52$& $0.51$ & $0.66$ \\
        &T2I-R1-1B &  $0.99_{\pm 0.01}$& $0.85_{\pm 0.03}$& $0.50_{\pm 0.03}$& $0.86_{\pm 0.01}$& $0.64_{\pm 0.05}$ & $0.63_{\pm 0.02}$ & $0.75_{\pm 0.01}$ \\
        &\rowcolor{verylightgray}
\textbf{IRIS}-1B &  $0.99_{\pm 0.01}$& $0.85_{\pm 0.01}$& $0.41_{\pm 0.03}$& $0.88_{\pm 0.02}$& $0.66_{\pm 0.04}$ & $0.51_{\pm 0.03}$ & $0.72_{\pm 0.01}$ \\
&Janus-Pro-7B &  $0.98$& $0.89$& $0.49$& $0.89$& $0.69$& $0.62$ & $0.76$ \\
&T2I-R1-7B& $1.00_{\pm 0.01}$& $0.91_{\pm 0.03}$& $0.55_{\pm 0.03}$& $0.91_{\pm 0.01}$& $0.69_{\pm 0.04}$& $0.62_{\pm 0.03}$& $0.78_{\pm 0.01}$
\\
&\rowcolor{verylightgray}
\textbf{IRIS}-7B (Ours) & $0.99_{\pm 0.01}$& $0.91_{\pm 0.03}$& $0.52_{\pm 0.06}$& $0.88_{\pm 0.03}$& $0.73_{\pm 0.05}$& $0.61_{\pm 0.03}$& $0.77_{\pm 0.03}$
\\
\bottomrule
    \end{NiceTabular}
}
\label{tab:benchmark_geneval}
\end{table}

\subsection{Additional ablation studies}\label{app:additional_ablation}
\begin{figure}[!h]
  \centering
  \includegraphics[width=0.4\linewidth]{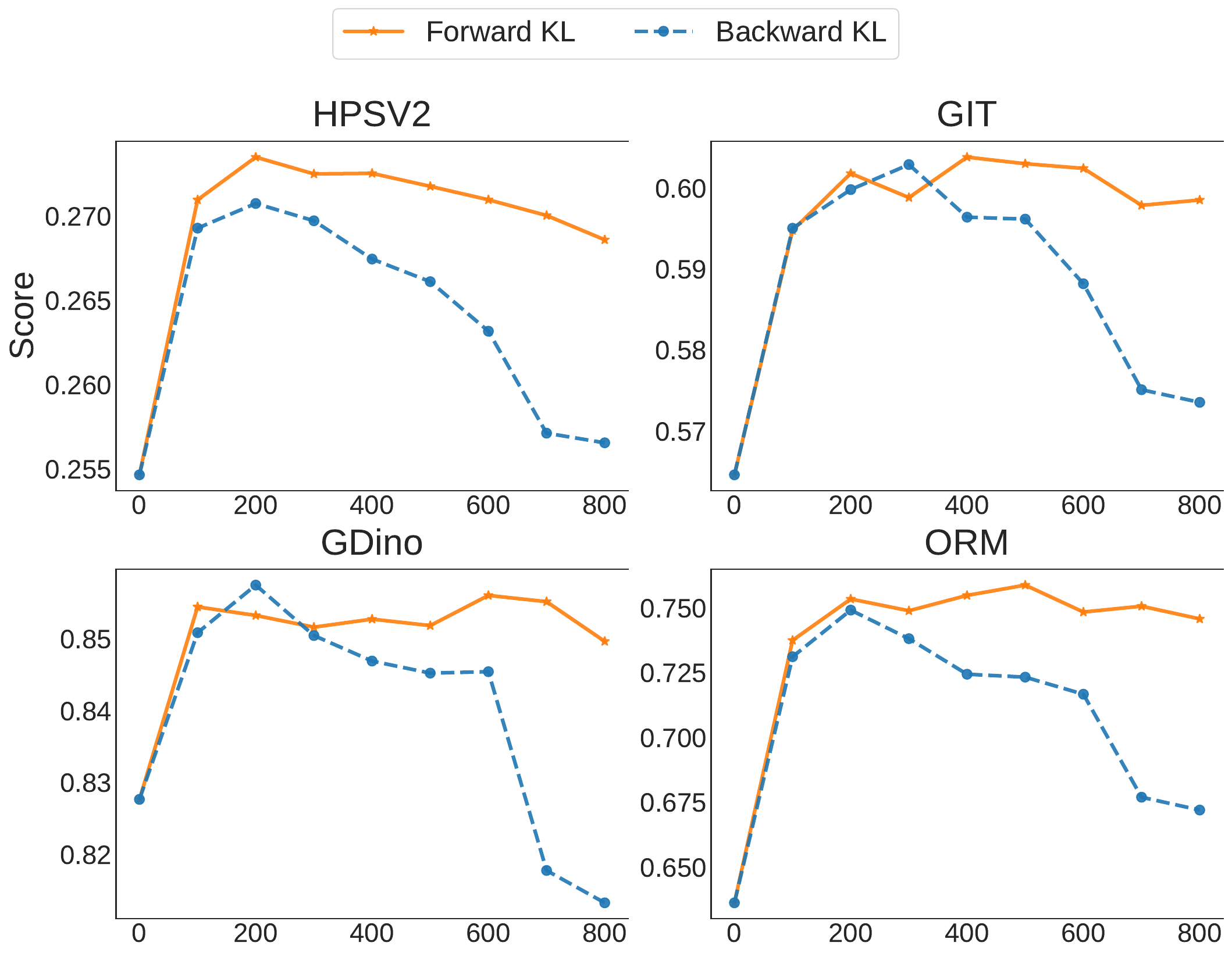} 
  \caption{\textbf{Ablation}: forward KL (ours) outperforms backward KL.}
  \label{fig:ablation_entropy_vs_sc}
\end{figure}

\paragraph{Forward or backward KL} We consider the backward KL divergence formulation of NSC reward, 
\begin{align*}
\operatorname{NSC}^\prime(o_t|q,\Vec{o}_{<t}) := - {\rm KL}(\pi_{\theta}(o_{t}|q,\Vec{o}_{<t})\|U)
={\rm Entropy}(\pi_{\theta}(\Vec{o}_{t}|q,o_{<t}))-\log|U|\,,
\end{align*}
where $|U|$ is the vocabulary size. Compared with minimizing the forward KL divergence with respect to the uniform distribution, minimizing the backward KL divergence is equivalent to maximizing the entropy. In \cref{fig:ablation_entropy_vs_sc}, we show that backward KL divergence formulation is subpar to the forward counterpart, which is consistent with previous findings that self-certainty behaves better than entropy in text models~\citep{zhao2025learning,kang2025scalable}. 

\subsection{CoT examples}\label{app:subsec:failed_cases} \cref{fig:text_prompts} illustrates that semantic CoTs generated by the model trained with intrinsic rewards at the step 0 (the base model), step 400, and step 800. We find that RL finetuning on intrinsic rewards can incentivize the emergence of descriptive CoTs. 

\begin{figure}[!htbp]
\centering
\begin{minipage}{\linewidth}
\begin{tcolorbox}[
    colback=blue!5!white,
    colframe=blue!75!black,
    title=Single object, 
    fonttitle=\bfseries,
    sharp corners,
    boxsep=5pt,
    left=12pt, 
    drop shadow
]
\begin{description}
  \item[Prompt:] a photo of a computer keyboard

  \item[Step 0:] Here is a 50-word, concise visualization of the prompt: A laptop keyboard in black and white.

  \item[Step 400:] A computer keyboard with a white keyboard, green keys labeled 'A' through 'Z', blue arrow keys with numbers from '0' to '9', and functional keys arranged in a standard QWERTY layout.

  \item[Step 800:] Here is a visualization of a photo of a computer keyboard: A photo of a computer keyboard in the foreground, a white screen in the background, and a black keys with numeric keys on the right-hand side. The keys have a black background, and the numbers on them are blue with thick black borders. The screen is white with a faint grid pattern in the middle. Space is arranged uniformly around the keyboard, giving the image a clean and organized appearance. The keys on the keyboard are arranged in a logical pattern of function with the arrow keys highlighted. No extraneous text is present in this image.
\end{description}
\end{tcolorbox}

\begin{tcolorbox}[
    colback=blue!5!white,
    colframe=blue!75!black,
    title=Color attribution,
    fonttitle=\bfseries,
    sharp corners,
    boxsep=5pt,
    left=12pt, 
    drop shadow
]
\begin{description}
  \item[Prompt:] a photo of an orange traffic light and a white toilet

  \item[Step 0:]  [Image of an orange traffic light with "STOP" written on it]
[Image of a white toilet with the label "TOILET" written on it]

  \item[Step 400:] A close-up photo showcasing an orange traffic light with a white toilet in the background. The orange traffic light features a reflective surface with a glossy sheen, the toilet has a smooth, ceramic texture with a typical white finish, placed in front of a wooden fence.

  \item[Step 800:] The image shows an orange traffic light and a white toilet, with the traffic light at the center and the toilet positioned to the top right side of the image. The traffic light is rectangular with orange stripes and a lit bulb, and the toilet is round with white walls, a white seat, and a large opening. The contrast between the bright orange and white colors creates a striking visual, with the traffic light standing out as the primary object while adding a touch of humor to the mundane scene.
  
\end{description}
\end{tcolorbox}

\begin{tcolorbox}[
    colback=blue!5!white,
    colframe=blue!75!black,
    title=Position,
    fonttitle=\bfseries,
    sharp corners,
    boxsep=5pt,
    left=12pt, 
    drop shadow
]
\begin{description}
  \item[Prompt:] a photo of a couch below a potted plant

  \item[Step 0:] Here is a 50-word, concise visualization of the prompt: A photo of a couch below a potted plant with a gradient background transitioning from green to blue.

  \item[Step 400:] A cozy-looking couch with light brown and beige tones is adorned below a small green potted plant with smooth, rounded leaves. The couch features a cushiony fabric, and the plant's vibrant green complements its natural appearance. The scene is vibrant and lively, set in a bright, sunlit room. 

  \item[Step 800:] The image shows a couch and a potted plant positioned underneath one another, both in a living room setting. The color of both elements is green, the couch is light brown in color with a wood texture, and the plant has a vibrant green color without any visible leaves. Shape-wise, the couch is rectangular, while the potted plant is round with a leafy texture. There is a faint background of neutral colors, emphasizing the contrast between the couch and the plant. The relationship between the elements is that the plant is in an angled position in front of the couch, complementing the shape and placement of the sitting furniture.
  
\end{description}
\end{tcolorbox}

\end{minipage}
\caption{Examples of semantic CoTs given the prompts in the three categories: \textit{single object}, \textit{color attribution} and \textit{position}.}
\label{fig:text_prompts}
\end{figure}

\subsection{Image generation examples}\label{app:subsec:visualized_results}
\begin{figure}[!htbp]
    \centering
    \begin{subfigure}{\linewidth}
        \centering
        \includegraphics[width=0.55\linewidth]{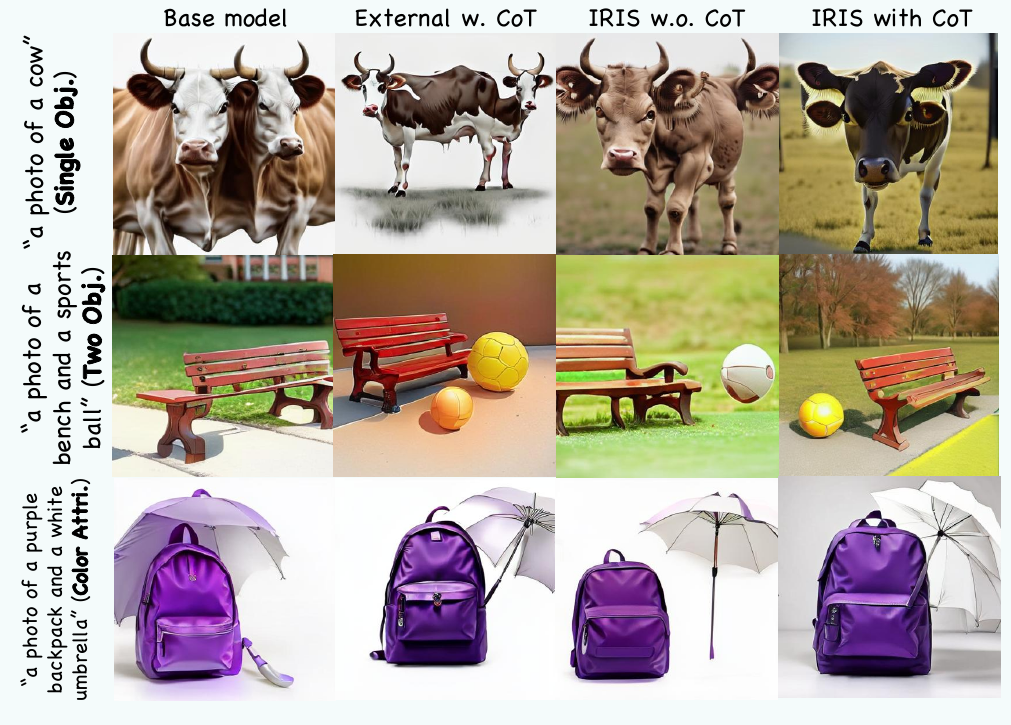}
        \subcaption{GenEval.}
    \end{subfigure}
    \begin{subfigure}{\linewidth}
        \centering
        \includegraphics[width=0.55\linewidth]{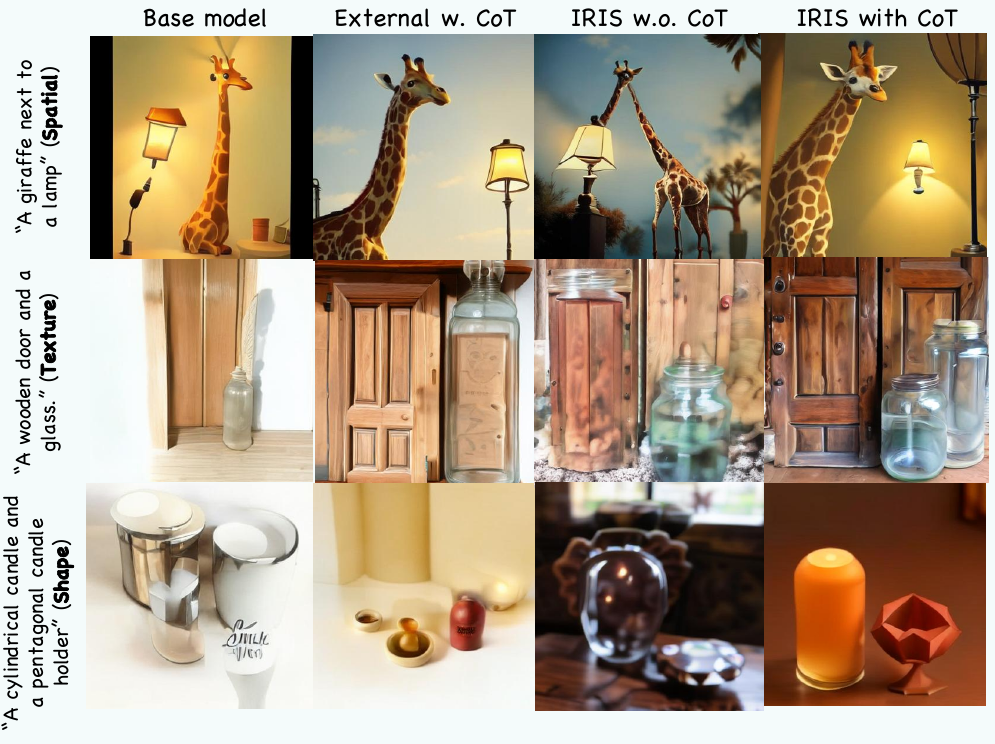}
        \subcaption{T2I-CompBench.}
    \end{subfigure}
    \begin{subfigure}{\linewidth}
        \centering
        \includegraphics[width=0.55\linewidth]{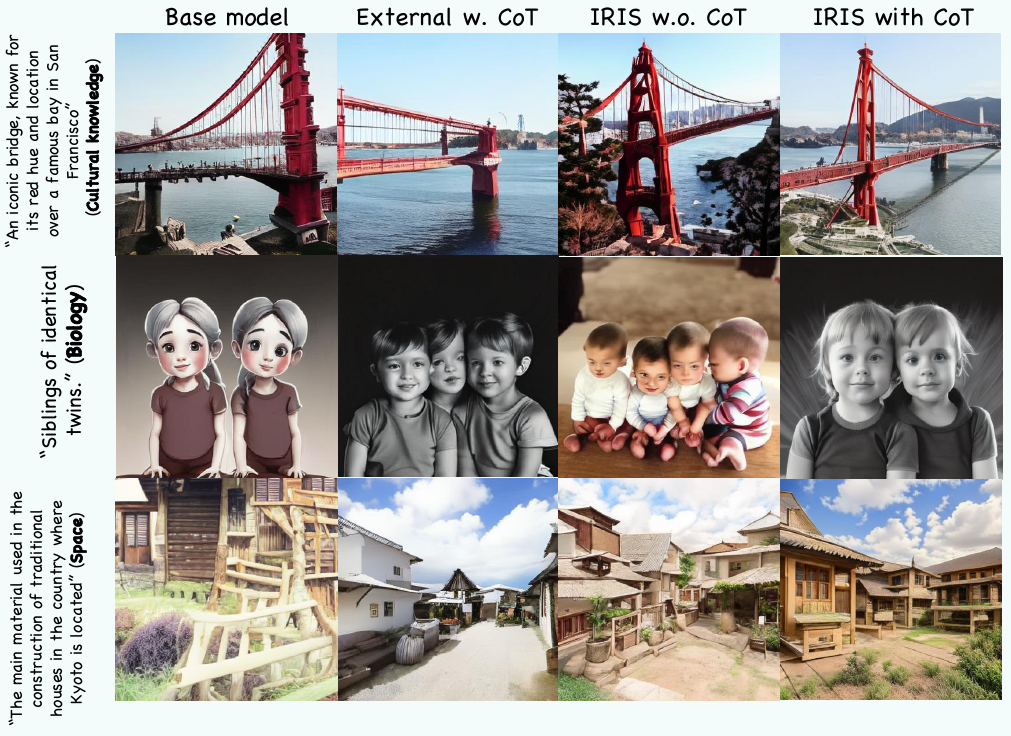}
        \subcaption{WISE.}
    \end{subfigure}
    \label{fig:examples_compwise}
    \vspace{-1.5em}
    \caption{Examples of generated images.}
\end{figure}

In \cref{fig:examples_compwise}, we present some examples. The base model is Janus-Pro 1B. We see that IRIS with CoTs (default in the main paper) can generate the most vivid and descriptive images.

\subsection{Image generation prompts}
Here, we present the prompt to generate the images on Janus Pro models in training and evaluation. 
\begin{tcolorbox}[title=Image generation prompts ,colback=blue!5!white, colframe=blue!75!black, fonttitle=\bfseries, sharp corners, boxsep=5pt, left=12pt, drop shadow]
\begin{lstlisting}
You are asked to generate an image based on this prompt: "{}"
Provide a brief, precise visualization of all elements in the prompt. Your description should:
1. Include every object mentioned in the prompt
2. Specify visual attributes (color, number, shape, texture) if specified in the prompt
3. Clarify relationships (e.g., spatial) between objects if specified in the prompt
4. Be concise (50 words or less)
5. Focus only on what's explicitly stated in the prompt
6. Do not elaborate beyond the attributes or relationships specified in the prompt
Do not miss objects. Output your visualization directly without explanation: 
\end{lstlisting}
\end{tcolorbox}




\end{document}